%% file: main.tex
\documentclass[10pt,twocolumn,letterpaper]{article}

\usepackage{cvpr}              %
\usepackage{paralist}

\input{preamble}

\definecolor{cvprblue}{rgb}{0.21,0.49,0.74}
\definecolor{regressionloss}{rgb}{.706, .827, .718}
\definecolor{distrmatchgrad}{rgb}{.929, .110, .137}
\definecolor{diffusionloss}{rgb}{.533, .616, .816}

\usepackage[pagebackref,breaklinks,colorlinks,citecolor=cvprblue]{hyperref}
\usepackage[capitalize]{cleveref}
\crefname{section}{Sec.}{Secs.}
\Crefname{section}{Section}{Sections}
\Crefname{table}{Table}{Tables}
\crefname{table}{Tab.}{Tabs.}

\def\paperID{1534} %
\def\confName{CVPR}
\def\confYear{2024}
\def\method{DMD\xspace}

\title{One-step Diffusion with Distribution Matching Distillation}

\author{Tianwei Yin$^{1}$ \hspace{10mm} Michaël Gharbi$^{2}$ \hspace{10mm} Richard Zhang$^{2}$ \hspace{10mm} Eli Shechtman$^{2}$ \\ 
Frédo Durand$^{1}$ \hspace{10mm} William T. Freeman$^{1}$ \hspace{10mm} Taesung Park$^{2}$ \\ \vspace{-3mm} \\
$^{1}$Massachusetts Institute of Technology \hspace{14mm} $^{2}$Adobe Research \\
\centerline{\href{https://tianweiy.github.io/dmd/}{https://tianweiy.github.io/dmd/}}
}

\begin{document}

\twocolumn[{%
 \renewcommand\twocolumn[1][]{#1}%
 \maketitle
 \vspace{-6mm}
 \centering
 \includegraphics[width=\textwidth]{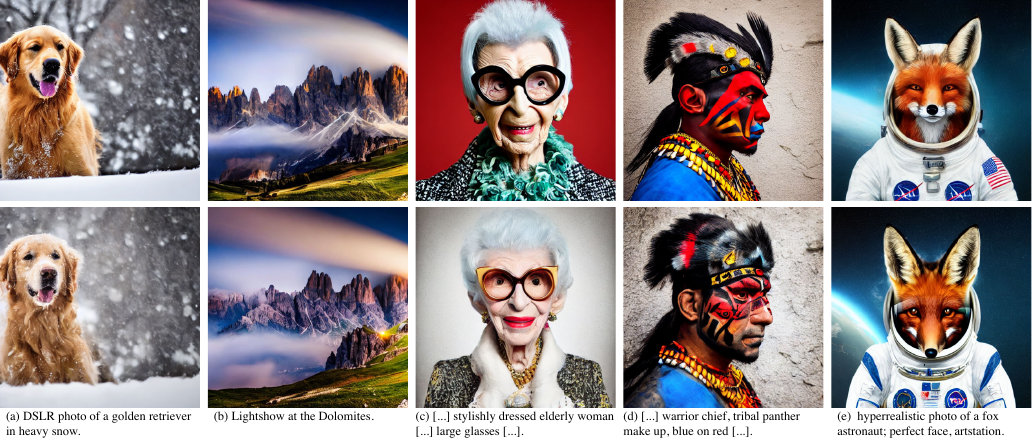}
 \vspace{-8mm}
 \captionof{figure}{ 
    {\bf Which is which?}
    Among these images, some were generated with baseline {\bf Stable Diffusion (SD)~\cite{rombach2022high} (2590ms each)}, the others with our {\bf Diffusion Matching Distillation (DMD) (90ms each)}.
    Can you tell which is which?
    Answers in the footnote\protect\footnotemark.
    (Non-abbreviated prompts in Appendix~\ref{sec:fig1_prompts}.) Our one-step text-to-image generators provide
    quality rivaling expensive diffusion models. %
    \label{fig:teaser}
   }
   \vspace{2mm}
}]

\maketitle

\input{00_abstract}

\footnotetext{
\begin{minipage}[t] {3in}
{\bf Ours (left to right):}
bottom, top, bottom, bottom, top.
\end{minipage}%
}%

\vspace{-5mm}
\section{Introduction}
\input{01_intro}

\section{Related Work}
\input{02_related}

\section{Distribution Matching Distillation}

\input{04_method}

\section{Experiments}
\input{05_experiments}

\section{Limitations}
While our results are promising, a slight quality discrepancy persists between our one-step model and finer discretizations of the diffusion sampling path, such as those with 100 or 1000 neural network evaluations. 
Additionally, our framework fine-tunes the weights of both the fake score function and the generator, leading to significant memory usage during training. 
Techniques such as LORA offer potential solutions for addressing this issue.

\section*{Acknowledgements}

This work was started while TY was an intern at Adobe Research.
We are grateful for insightful discussions with Yilun Xu, Guangxuan Xiao, and Minguk Kang. 
This work is supported, in part by NSF grants 2105819,  1955864, and 2019786 (IAIFI), by the Singapore DSTA under DST00OECI20300823 (New Representations for Vision), as well as by funding from GIST and Amazon.

\clearpage 

{\small
\bibliographystyle{ieee_fullname}
\bibliography{main}
}

\clearpage 
\appendix 

\noindent{\Large\bf Appendix}

\section{Qualitative Speed Comparison}
In the accompanying video material, we present a qualitative speed comparison between our one-step generator and the original stable diffusion model.
Our one-step generator achieves comparable image quality with the Stable Diffusion model while being around $30\times$ faster.

\section{Implementation Details}
\label{sec:implementation}

For a comprehensive understanding, we include the implementation specifics for constructing the KL loss for the generator~$G$ in Algorithm~\ref{alg:generator-grad} and training the fake score estimator parameterized by $\mu_\text{fake}$ in Algorithm~\ref{alg:fake-grad}.

\begin{algorithm}[t]
\caption{distributionMatchingLoss}
\label{alg:generator-grad}
\definecolor{codeblue}{rgb}{0.25,0.5,0.5}
\lstset{
  backgroundcolor=\color{white},
  basicstyle=\fontsize{7.2pt}{7.2pt}\ttfamily\selectfont,
  columns=fullflexible,
  breaklines=true,
  captionpos=b,
  commentstyle=\fontsize{7.2pt}{7.2pt}\color{codeblue},
  keywordstyle=\fontsize{7.2pt}{7.2pt},
}
\begin{lstlisting}[language=python]
# mu_real, mu_fake: denoising networks for real and fake distribution
# x: fake sample generated by our one-step generator
# min_dm_step, max_dm_step: timestep intervals for computing distribution matching loss 
# bs: batch size

# random timesteps 
timestep = randint(min_dm_step, max_dm_step, [bs])
noise = randn_like(x)

# Diffuse generated sample by injecting noise
# e.g. noise_x = x + noise * sigma_t (EDM)
noisy_x = forward_diffusion(x, noise, timestep) 

# denoise using real and fake denoiser
with_grad_disabled():
    pred_fake_image = mu_fake(noisy_x, timestep)
    pred_real_image = mu_real(noisy_x, timestep)

# The weighting_factor diverges slightly from our
# paper's equation, adapting to accomodate the mean
# prediction scheme we use here. 
weighting_factor = abs(x - pred_real_image).mean(
    dim=[1, 2, 3], keepdim=True)  
grad = (pred_fake_image - pred_real_image) / weighting_factor

# the loss that would enforce above grad
loss = 0.5 * mse_loss(x, stopgrad(x - grad))  

\end{lstlisting}
\end{algorithm}

\begin{algorithm}[t]
\caption{denoisingLoss}
\label{alg:fake-grad}
\definecolor{codeblue}{rgb}{0.25,0.5,0.5}
\lstset{
  backgroundcolor=\color{white},
  basicstyle=\fontsize{7.2pt}{7.2pt}\ttfamily\selectfont,
  columns=fullflexible,
  breaklines=true,
  captionpos=b,
  commentstyle=\fontsize{7.2pt}{7.2pt}\color{codeblue},
  keywordstyle=\fontsize{7.2pt}{7.2pt},
}
\begin{lstlisting}[language=python]
# pred_fake_image: denoised output by mu_fake on x_t
# x: fake sample generated by our one-step generator
# weight: weighting strategy(SNR+1/0.5^2 for EDM, SNR for SDv1.5)
loss = mean(weight*(pred_fake_image - x)**2)
\end{lstlisting}
\end{algorithm}

\subsection{CIFAR-10}
\label{sec:cifar10}
We distill our one-step generator from EDM~\cite{karras2022elucidating} pretrained models, specifically utilizing ``edm-cifar10-32x32-cond-vp" for class-conditional training and ``edm-cifar10-32x32-uncond-vp" for unconditional training.
We use $\sigma_\text{min}=0.002$ and $\sigma_\text{max} =80$ and discretize the noise schedules into 1000 bins\footnote{\url{https://github.com/openai/consistency_models/blob/main/cm/karras_diffusion.py\#L422}}.
To create our distillation dataset, we generate 100,000 noise-image pairs for class-conditional training and 500,000 for unconditional training. 
This process utilizes the deterministic Heun sampler~(with $S_\text{churn}=0$) over 18 steps~\cite{karras2022elucidating}.
For the training phase, we use the AdamW optimizer~\cite{loshchilov2017decoupled}, setting the learning rate at 5e-5, weight decay to 0.01, and beta parameters to (0.9, 0.999).
We use a learning rate warmup of 500 steps. 
The model training is conducted across 7 GPUs, achieving a total batch size of 392. Concurrently, we sample an equivalent number of noise-image pairs from the distillation dataset to calculate the regression loss.
Following Song \textit{et al.}~\cite{song2023consistency}, we incorporate the LPIPS loss using a VGG backbone from the PIQ library~\cite{kastryulin2022piq}.
Prior to input into the LPIPS network, images are upscaled to a resolution of 224×224 using bilinear upsampling. The regression loss is weighted at 0.25~($\lambda_\text{reg}=0.25$) for class-conditional training and at 0.5~($\lambda_\text{reg}=0.5$) for unconditional training.
The weights for the distribution matching loss and fake score denoising loss are both set to 1.
We train the model for 300,000 iterations and use a gradient clipping with a L2 norm of 10.
The dropout is disabled for all networks following consistency model~\cite{song2023consistency}.

\subsection{ImageNet-64$\times$64}
\label{sec:imagenet}
We distill our one-step generator from EDM~\cite{karras2022elucidating} pretrained models, specifically utilizing ``edm-imagenet-64x64-cond-adm" for class-conditional training.
We use a $\sigma_\text{min}=0.002$ and $\sigma_\text{max} =80$ and discretize the noise schedules into 1000 bins.
Initially, we prepare a distillation dataset by generating 25,000 noise-image pairs using the deterministic Heun sampler (with $S_\text{churn}=0$) over 256 steps~\cite{karras2022elucidating}.
For the training phase, we use the AdamW optimizer~\cite{loshchilov2017decoupled}, setting the learning rate at 2e-6, weight decay to 0.01, and beta parameters to (0.9, 0.999).
We use a learning rate warmup of 500 steps. 
The model training is conducted across 7 GPUs, achieving a total batch size of 336. 
Concurrently, we sample an equivalent number of noise-image pairs from the distillation dataset to calculate the regression loss.
Following Song \textit{et al.}~\cite{song2023consistency}, we incorporate the LPIPS loss using a VGG backbone from the PIQ library~\cite{kastryulin2022piq}.
Prior to input into the LPIPS network, images are upscaled to a resolution of 224$\times$224 using bilinear upsampling. The regression loss is weighted at 0.25~($\lambda_\text{reg}=0.25$), and the weights for the distribution matching loss and fake score denoising loss are both set to 1.
We train the models for 350,000 iterations.
We use mixed-precision training and a gradient clipping with a L2 norm of 10.
The dropout is disabled for all networks following consistency model~\cite{song2023consistency}.

\subsection{LAION-Aesthetic 6.25+}
We distill our one-step generator from Stable Diffusion v1.5~\cite{rombach2022high}.
We use the LAION-Aesthetic 6.25+~\cite{schuhmann2022laion} dataset, which contains around 3 million images. 
Initially, we prepare a distillation dataset by generating 500,000 noise-image pairs using the deterministic PNMS sampler~\cite{liu2022pseudo} over 50 steps with a guidance scale of 3.
Each pair corresponds to one of the first 500,000 prompts of LAION-Aesthetic 6.25+.
For the training phase, we use the AdamW optimizer~\cite{loshchilov2017decoupled}, setting the learning rate at 1e-5, weight decay to 0.01, and beta parameters to (0.9, 0.999).
We use a learning rate warmup of 500 steps. 
The model training is conducted across 72 GPUs, achieving a total batch size of 2304. 
Simultaneously, noise-image pairs from the distillation dataset are sampled to compute the regression loss, with a total batch size of 1152.
Given the memory-intensive nature of decoding generated latents into images using the VAE for regression loss computation, we opt for a smaller VAE network~\cite{taesd} for decoding. 
Following Song \textit{et al.}~\cite{song2023consistency}, we incorporate the LPIPS loss using a VGG backbone from the PIQ library~\cite{kastryulin2022piq}.
The regression loss is weighted at 0.25~($\lambda_\text{reg}=0.25$), and the weights for the distribution matching loss and fake score denoising loss are both set to 1.
We train the model for 20,000 iterations. 
To optimize GPU memory usage, we implement gradient checkpointing~\cite{chen2016training} and mixed-precision training.
We also apply a gradient clipping with a L2 norm of 10.

\subsection{LAION-Aesthetic 6+}
We distill our one-step generator from Stable Diffusion v1.5~\cite{rombach2022high}.
We use the LAION-Aesthetic 6+~\cite{schuhmann2022laion} dataset, comprising approximately 12 million images. To prepare the distillation dataset, we generate 12,000,000 noise-image pairs using the deterministic PNMS sampler~\cite{liu2022pseudo} over 50 steps with a guidance scale of 8.
Each pair corresponds to a prompt from the LAION-Aesthetic 6+ dataset.
For training, we utilize the AdamW optimizer~\cite{loshchilov2017decoupled}, setting the learning rate at 1e-5, weight decay to 0.01, and beta parameters to (0.9, 0.999).
We use a learning rate warmup of 500 steps. 
To optimize GPU memory usage, we implement gradient checkpointing~\cite{chen2016training} and mixed-precision training.
We also apply a gradient clipping with a L2 norm of 10.
The training takes two weeks on approximately 80 A100 GPUs. 
During this period, we made adjustments to the distillation dataset size, the regression loss weight, the type of VAE decoder, and the maximum timestep for the distribution matching loss computation.
A comprehensive training log is provided in Table~\ref{table:laion6}.
We note that this training schedule, constrained by time and computational resources, may not be the most efficient or optimal.

\begin{table}[h]
\centering
\setlength\tabcolsep{4.5pt} 
\resizebox{\linewidth}{!}{
\begin{tabular}{|ccccccccc|}
\hline
Version & \#Reg. Pair & Reg. Weight & Max DM Step & VAE-Type & DM BS & Reg. BS & Cumulative Iter. & FID \\
\hline
V1 & 2.5M & 0.1 & 980 & Small & 32 & 16 & 5400 & 23.88 \\
\hline
V2 & 2.5M & 0.5 & 980 & Small & 32 & 16 & 8600 & 18.21 \\
\hline
V3 & 2.5M & 1 & 980 & Small & 32 & 16 & 21100 & 16.10 \\
\hline
V4 & 4M & 1 & 980 & Small & 32 & 16 & 56300 & 16.86 \\
\hline
V5 & 6M & 1 & 980 & Small & 32 & 16 & 60100 & 16.94 \\
\hline
V6 & 9M & 1 & 980 & Small & 32 & 16 & 68000 & 16.76 \\
\hline
V7 & 12M & 1 & 980 & Small & 32 & 16 & 74000 & 16.80 \\
\hline
V8 & 12M & 1 & 500 & Small & 32 & 16 & 80000 & 15.61 \\
\hline
V9 & 12M & 1 & 500 & Large & 16 & 4 & 127000 & 15.33 \\
\hline
V10 & 12M & 0.75 & 500 & Large & 16 & 4 & 149500 & 15.51 \\
\hline
V11 & 12M & 0.5 & 500 & Large & 16 & 4 & 162500 & 15.05 \\
\hline
V12 & 12M & 0.25 & 500 & Large & 16 & 4 & 165000 & 14.93 \\
\hline
\end{tabular}
}
\caption{\label{table:laion6} Training Logs for the LAION-Aesthetic 6+ Dataset: `Max DM step' denotes the highest timestep for noise injection in computing the distribution matching loss. ``VAE-Type small'' corresponds to the Tiny VAE decoder~\cite{taesd}, while ``VAE-Type large'' indicates the standard VAE decoder used in SDv1.5. ``DM BS'' denotes the batch size used for the distribution matching loss while ``Reg. BS'' represents the batch size used for the regression loss. 
}
\end{table}

\section{Baseline Details}

\subsection{w/o Distribution Matching Baseline}
This baseline adheres to the training settings outlined in Sections~\ref{sec:cifar10} and \ref{sec:imagenet}, with the distribution matching loss omitted.

\subsection{w/o Regression Loss Baseline}
Following the training protocols from Sections~\ref{sec:cifar10} and \ref{sec:imagenet}, this baseline excludes the regression loss. To prevent training divergence, the learning rate is adjusted to 1e-5.

\subsection{Text-to-Image Baselines}

We benchmark our approach against a variety of models, including the base diffusion model~\cite{rombach2022high}, fast diffusion solvers~\cite{zhao2023unipc, lu2022dpm++}, and few-step diffusion distillation baselines~\cite{luo2023latent, luo2023latentlora}.

\noindent 
\textbf{Stable Diffusion}
We employ the StableDiffusion v1.5 model available on huggingface\footnote{\url{https://huggingface.co/runwayml/stable-diffusion-v1-5}}, generating images with the PNMS sampler~\cite{liu2022pseudo} over 50 steps.

\noindent 
\textbf{Fast Diffusion Solvers} 
We use the UniPC~\cite{zhao2023unipc} and DPMSolver++~\cite{lu2022dpm++} implementations from the diffusers library~\cite{von-platen-etal-2022-diffusers}, with all hyperparameters set to default values.

\noindent 
\textbf{LCM-LoRA}
We use the LCM-LoRA SDv1.5 checkpoints hosted on Hugging Face\footnote{\url{https://huggingface.co/latent-consistency/lcm-lora-sdv1-5}}.
As the model is pre-trained with guidance, we do not apply classifier-free guidance during inference.

\section{Evaluation Details}
\label{sec:evaluation}
For zero-shot evaluation on COCO, we employ the evaluation code from GigaGAN~\cite{kang2023scaling}\footnote{\url{https://github.com/mingukkang/GigaGAN/tree/main/evaluation}}. 
Specifically, we generate 30,000 images using random prompts from the MS-COCO2014 validation set.
We downsample the generated images from 512$\times$512 to 256$\times$256 using the PIL.Lanczos resizer.
These images are then compared with 40,504 real images from the same validation set to calculate the FID metric using the clean-fid~\cite{parmar2022aliased} library.
Additionally, we employ the OpenCLIP-G backbone to compute the CLIP score.
For ImageNet and CIFAR-10, we generate 50,000 images for each and calculate their FID using the EDM's evaluation code~\cite{karras2022elucidating}\footnote{\url{https://github.com/NVlabs/edm/blob/main/fid.py}}.

\section{CIFAR-10 Experiments}
\label{sec:cifar10_exp}
Following the setup outlined in Section~\ref{sec:cifar10}, we train our models on CIFAR-10 and conduct comparisons with other competing approaches. 
Table~\ref{table:cifar10} summarizes the results. 

\begin{table}[h]
\centering
\small 
\resizebox{\linewidth}{!}{
\begin{tabular}{llcc}
\toprule
\multirow{2}{*}{Family} & \multirow{2}{*}{Method} & \# Fwd & FID \\
& & Pass ($\downarrow$) & ($\downarrow$) \\
\midrule 
\multirow{11}{*}{\textbf{GAN}}
& BigGAN$^\dagger$ \cite{brock2018large} & 1 & 14.7 \\
& Diffusion GAN \cite{xiao2021tackling} & 1 & 14.6 \\
& Diffusion StyleGAN \cite{wang2022diffusion} & 1 & 3.19  \\
& AutoGAN \cite{gong2019autogan} & 1 & 12.4 \\
& E2GAN \cite{tian2020off} & 1 & 11.3 \\
& ViTGAN \cite{lee2021vitgan} & 1 & 6.66  \\
& TransGAN \cite{jiang2021transgan} & 1 & 9.26 \\
&  StylegGAN2 \cite{karras2020analyzing} & 1 & 6.96  \\
& StyleGAN2-ADA$^\dagger$ \cite{karras2020training} & 1 & 2.42 \\
& StyleGAN-XL$^\dagger$~\cite{sauer2022stylegan} & 1 & \textbf{1.85} \\ 
\midrule 

\multirow{5}{*}{\shortstack[l]{\textbf{Diffusion}\\ \textbf{+ Samplers}}} & DDIM \cite{song2020denoising}
& 10 & 8.23\\
& DPM-solver-2 \cite{lu2022dpm}
& 10 & 5.94\\
& DPM-solver-fast \cite{lu2022dpm}
& 10 & 4.70 \\
& 3-DEIS \cite{zheng2022fast}
& 10 & 4.17\\
& DPM-solver++ \cite{lu2022dpm++} & 10 & \textbf{2.91} \\
\midrule 
\multirow{11}{*}{\shortstack[l]{\textbf{Diffusion}\\ \textbf{+ Distillation}}} & Knowledge Distillation~\cite{luhman2021knowledge}
 & 1 & 9.36 \\
& DFNO \cite{zheng2022fast}
& 1 & 3.78 \\
& 1-Rectified Flow (+distill) \cite{liu2022flow}
 & 1 & 6.18 \\
& 2-Rectified Flow (+distill) \cite{liu2022flow}
 & 1 & 4.85\\
& 3-Rectified Flow (+distill) \cite{liu2022flow}
 & 1 & 5.21 \\
& Progressive Distillation \cite{salimans2022progressive} & 1 & 8.34  \\
& Meng et al.~\cite{meng2023distillation}$^\dagger$ & 1 & 5.98  \\
& Diff-Instruct~\cite{luo2023diff}$^\dagger$ & 1 & 4.19 \\
& Score Mismatching~\cite{ye2023score} & 1 & 8.10 \\
& TRACT \cite{berthelot2023tract} & 1 & 3.78 \\ 
& Consistency Model \cite{song2023consistency} & 1 & 3.55  \\
& \textbf{\method (Ours)} & 1 & 3.77  \\
& \textbf{\method-conditional (Ours)}$^\dagger$ & 1 & \textbf{2.66}  \\
\midrule 
\textbf{Diffusion} & EDM$^\dagger$ (Teacher) \cite{karras2022elucidating} & 35 & 1.84 \\
\bottomrule
\end{tabular}
}
\caption{
\label{table:cifar10}
Sample quality comparison on CIFAR-10. Baseline numbers are derived from Song et al.~\cite{song2023consistency}.
$^\dagger$Methods that use class-conditioning. 
}
\end{table}

\section{Derivation for Distribution Matching Gradient}
\label{sec:proof_dmd}

We present the derivation for Equation~\ref{eq:dm-grad} as follows:

\begin{equation}
\footnotesize 
\begin{aligned}
    \nabla_{\theta} D_{KL} &\simeq 
        \expect_{\substack{z, t, x, x_t
        }}
    \left[
        w_t
        \big(s_{\text{fake}}(x_t, t) - s_\text{real}(x_t, t)\big)
        \frac{\partial x_t}{\partial \theta}
        \right] \\
    &= 
        \expect_{\substack{z, t, x, x_t
        }}
    \left[
        w_t
        \big(s_{\text{fake}}(x_t, t) - s_\text{real}(x_t, t)\big)
        \frac{\partial x_t}{\partial G_\theta(z)}
         \frac{\partial G_\theta(z)}{\partial \theta}
        \right] \\
    &=
        \expect_{\substack{z, t, x, x_t
        }}
    \left[
        w_t
        \big(s_{\text{fake}}(x_t, t) - s_\text{real}(x_t, t)\big)
        \frac{\partial x_t}{\partial x}
         \frac{\partial G_\theta(z)}{\partial \theta}
        \right] \\  
&=
        \expect_{\substack{z, t, x, x_t
        }}
    \left[
        w_t \alpha_t
        \big(s_{\text{fake}}(x_t, t) - s_\text{real}(x_t, t)\big)
         \frac{dG}{d\theta}
        \right] \\  
\end{aligned}
\end{equation}

\section{Prompts for Figure~\ref{fig:teaser}}
\label{sec:fig1_prompts}
We use the following prompts for Figure~\ref{fig:teaser}. From left to right:
\begin{compactitem}
    \item A DSLR photo of a golden retriever in heavy snow.
    \item A Lightshow at the Dolomities.
    \item A professional portrait of a stylishly dressed elderly woman wearing very large glasses in the style of Iris Apfel, with highly detailed features.
    \item Medium shot side profile portrait photo of a warrior chief, sharp facial features, with tribal panther makeup in blue on red, looking away, serious but clear eyes, 50mm portrait, photography, hard rim lighting photography.
    \item A hyperrealistic photo of a fox astronaut; perfect face, artstation.
\end{compactitem}

\section{Equivalence of Noise and Data Prediction}
\label{sec:change_of_variable}

The noise prediction model $\epsilon(x_t, t)$ and data prediction model~$\mu(x_t, t)$ could be converted to each other according to the following rule~\cite{karras2022elucidating} 

\begin{equation}
\small 
    \begin{aligned}
    \mu(x_t, t) &= \frac{x_t - \sigma_t \epsilon(x_t, t)}{\alpha_t }, \quad
    \epsilon(x_t, t) = \frac{x_t - \alpha_t \mu(x_t, t)}{\sigma_t}.
    \end{aligned}
\end{equation}

\section{Further Analysis of the Regression Loss}
DMD utilizes a regression loss to stabilize training and mitigate mode collapse~(Sec.~\ref{sec:direct-distillation}). 
In our paper, we mainly adopt the LPIPS~\cite{zhang2018unreasonable} distance function, as it has been commonly adopted in prior works.
For further analysis, we experiment with a standard L2 distance to train our distilled model on the CIFAR-10 dataset. 
The model trained using L2 loss achieves an FID score of 2.78, compared to 2.66 with LPIPS, demonstrating the robustness of our method to different loss functions.

\section{More Qualitative Results}

\begin{figure*}[!h]
    \centering
    \includegraphics[width=\textwidth]{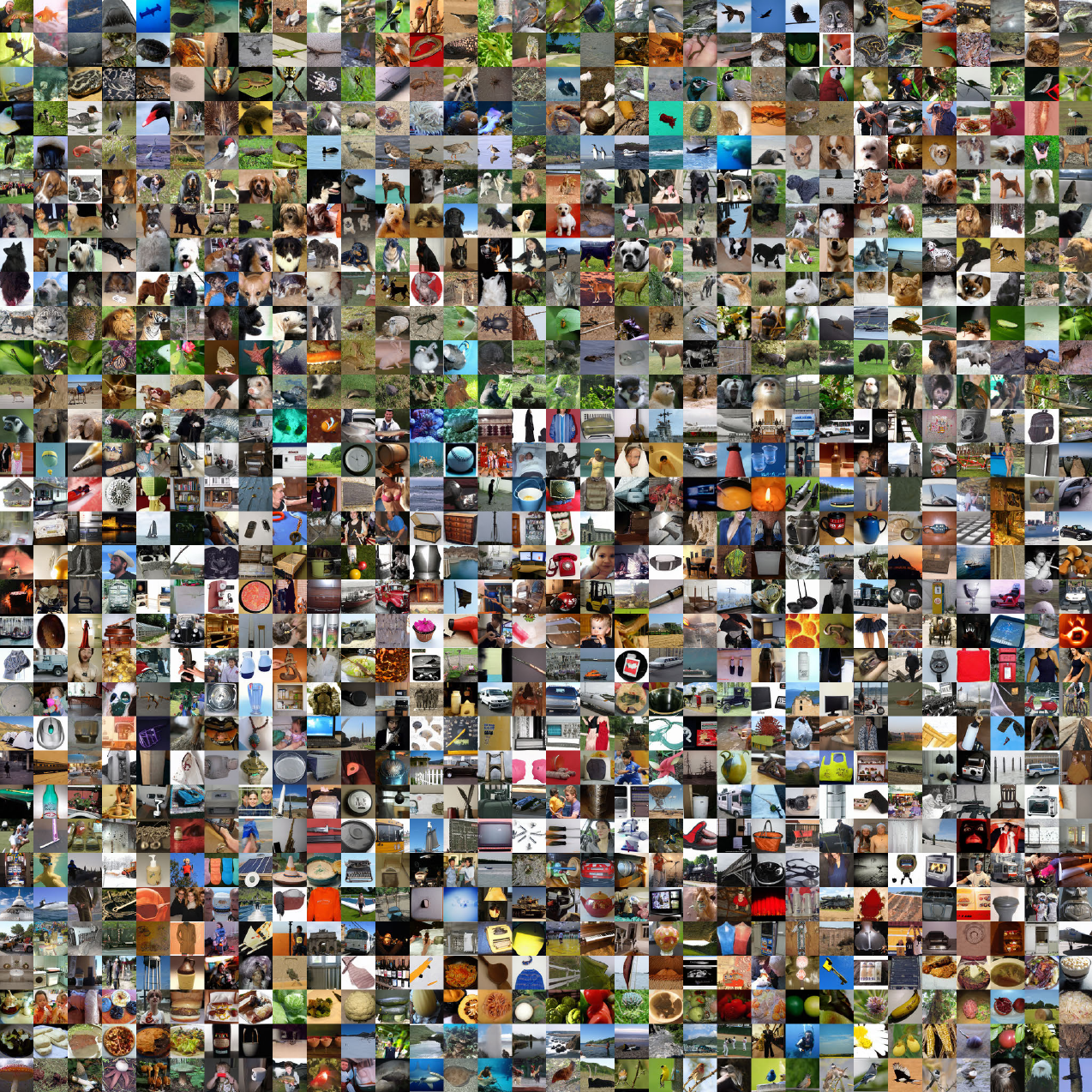}
    \caption{
        One-step samples from our class-conditional model on ImageNet~(FID=2.62).\label{fig:imagenet_cond}
        }
\end{figure*}

\begin{figure*}[!h]
    \centering
    \includegraphics[width=\textwidth]{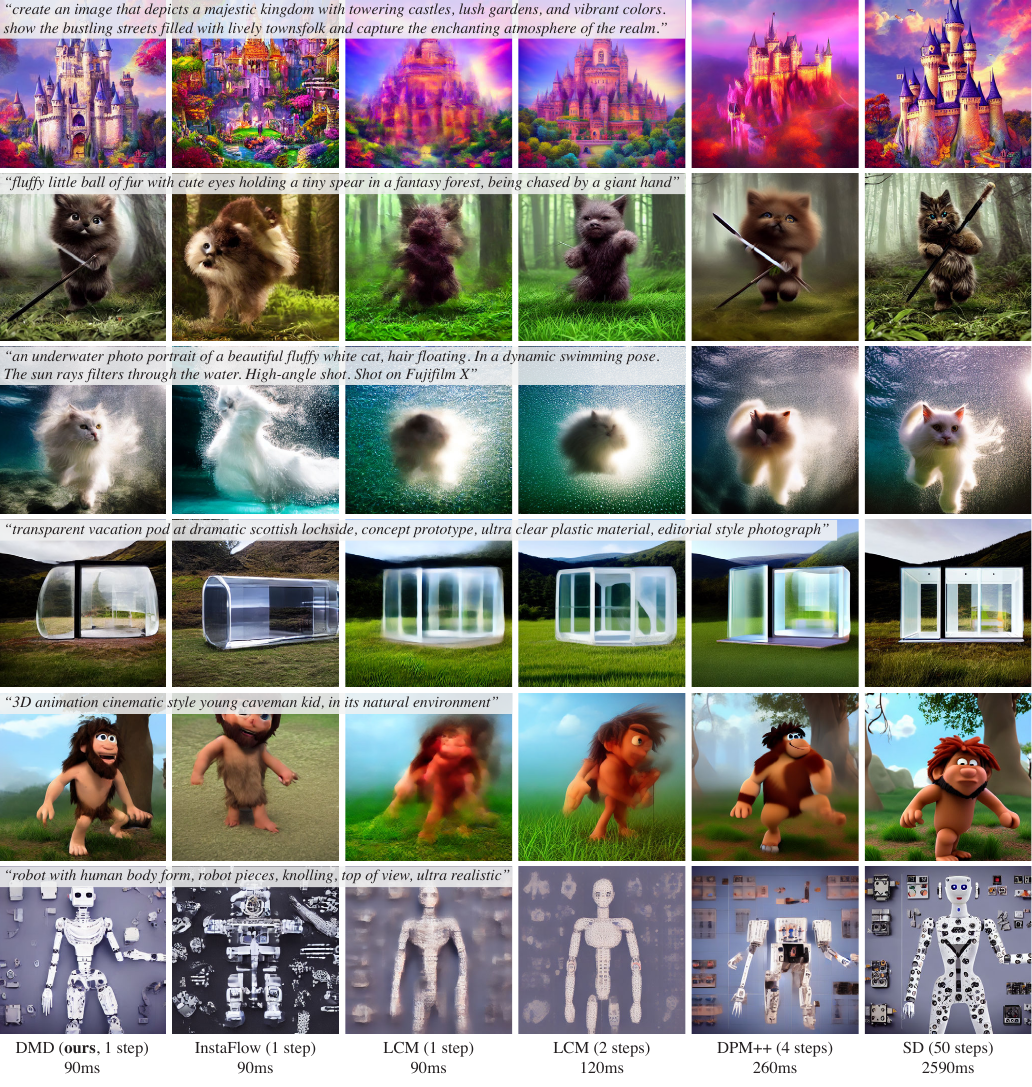}
    \caption{Starting from a pretrained diffusion model, here Stable Diffusion (right),
        our distribution matching distillation algorithm yields a model that can generate
        images with much higher quality (left)
        than previous few-steps generators (middle),
        with the same speed or faster.
        \label{fig:more_results}
        }
\end{figure*}

\begin{figure*}[!h]
    \centering
    \includegraphics[width=\textwidth]{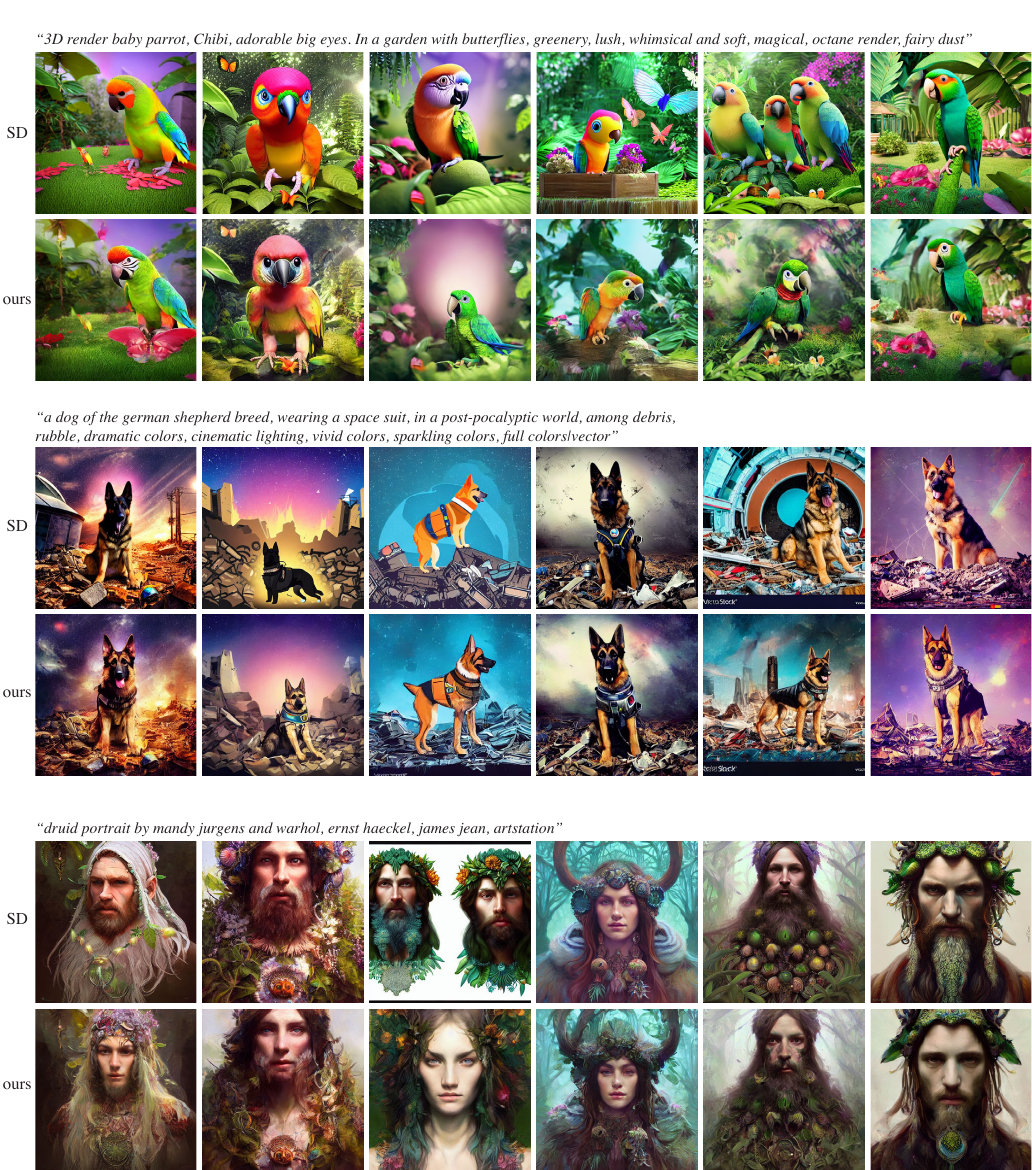}
    \caption{
        One-step samples from our LAION model. Our generator achieves comparable image quality with Stable Diffusion model at a speed $30\times$ faster.
        \label{fig:laion1}
        }
\end{figure*}

\begin{figure*}[!h]
    \centering
    \includegraphics[width=\textwidth]{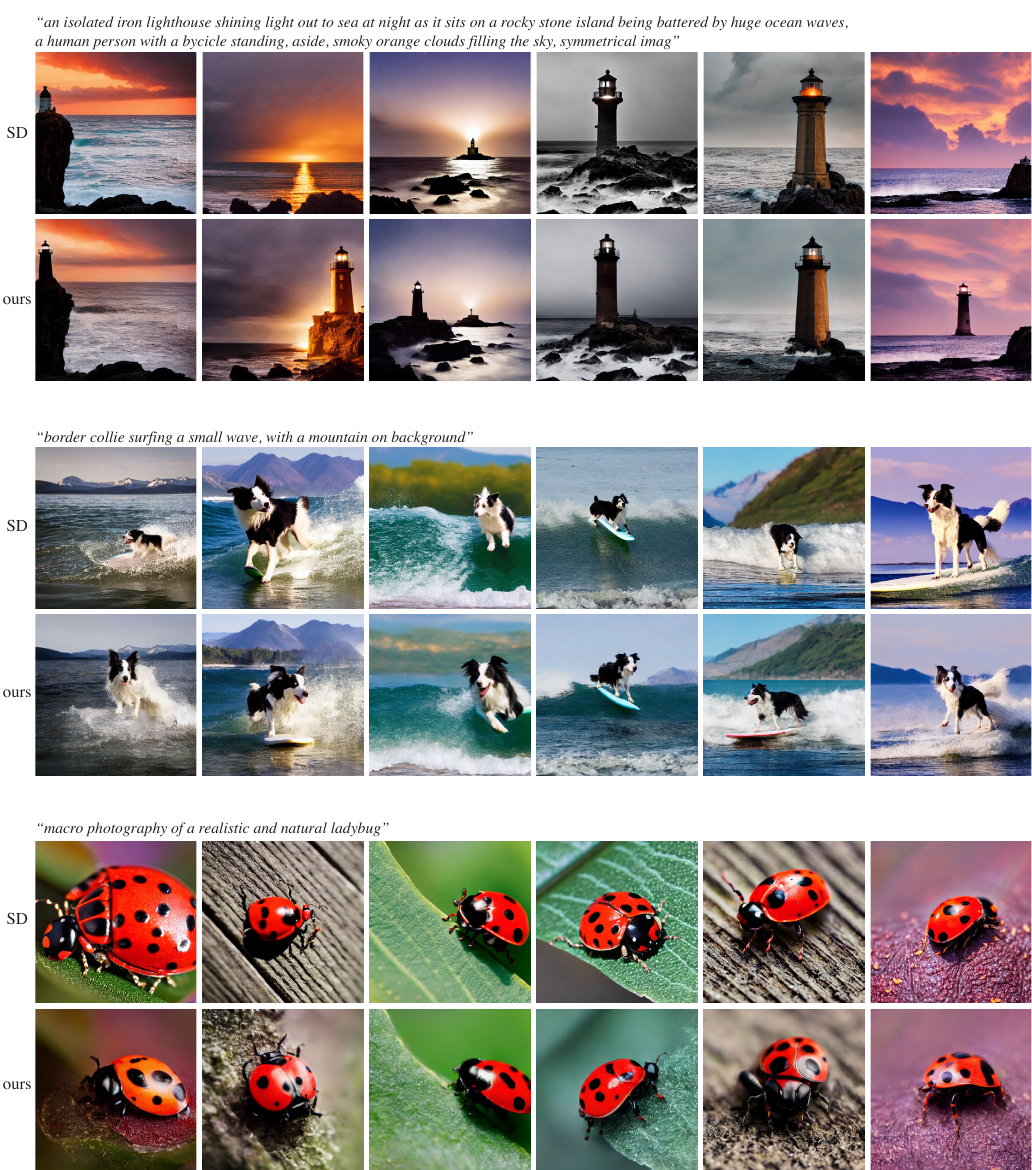}
    \caption{
        One-step samples from our LAION model. Our generator achieves comparable image quality with Stable Diffusion model at a speed $30\times$ faster.
        \label{fig:laion2}
        }
\end{figure*}

\begin{figure*}[!h]
    \centering
    \includegraphics[width=\textwidth]{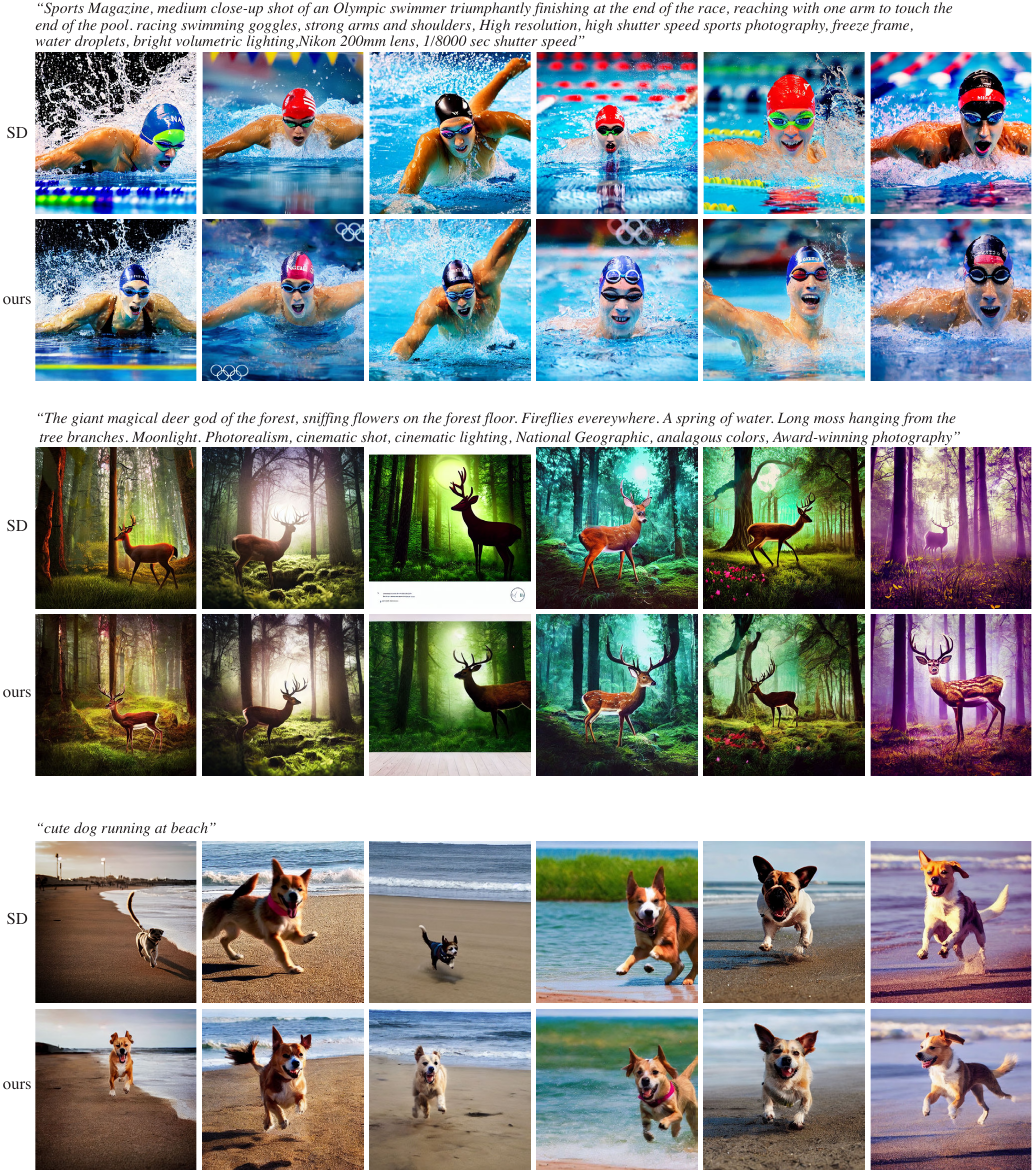}
    \caption{
        One-step samples from our LAION model. Our generator achieves comparable image quality with Stable Diffusion model at a speed $30\times$ faster.
        \label{fig:laion3}
        }
\end{figure*}

\begin{figure*}[!h]
    \centering
    \includegraphics[width=\textwidth]{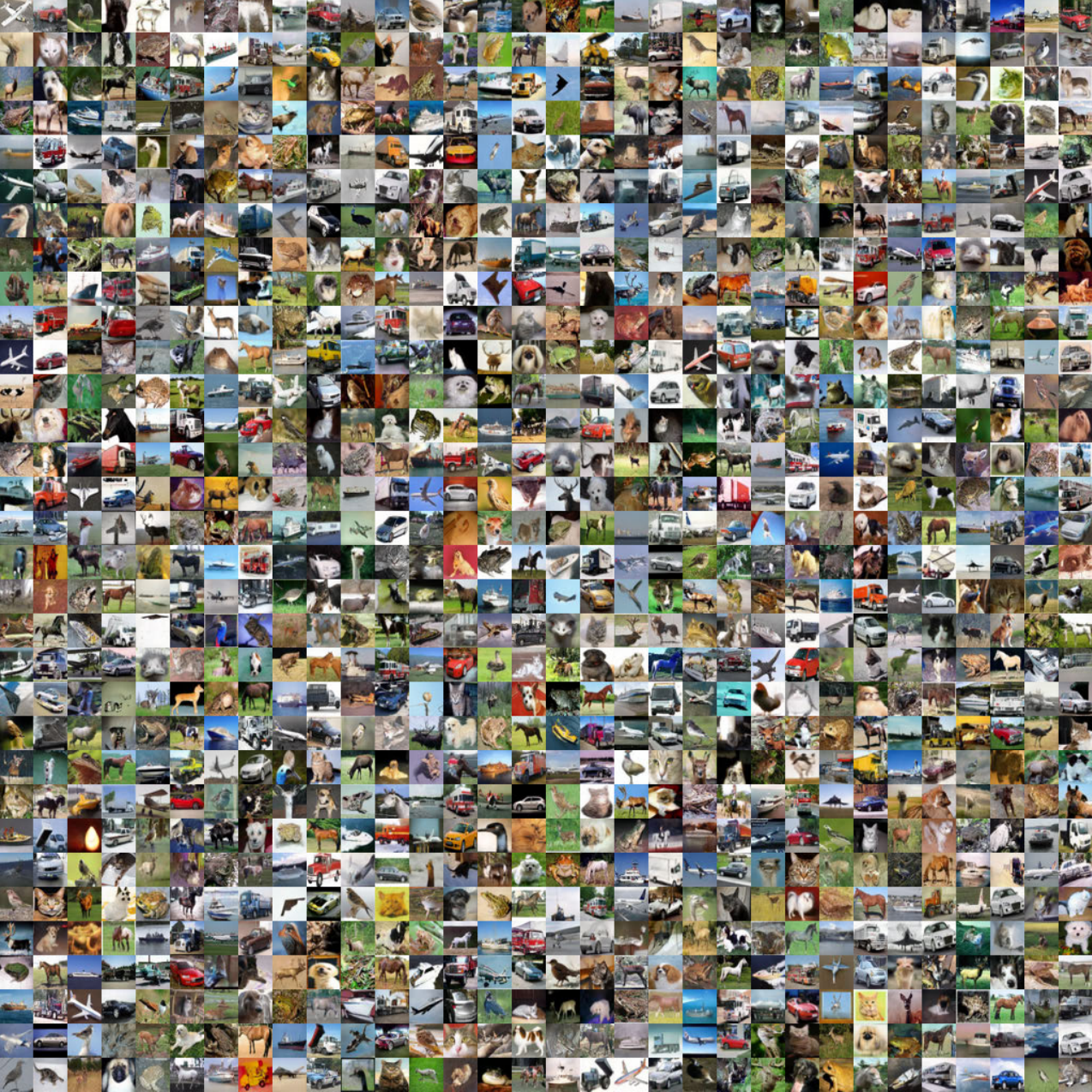}
    \caption{
        One-step samples from our class-conditional model on CIFAR-10~(FID=2.66).
        \label{fig:cifar10_cond}
        }
\end{figure*}

\begin{figure*}[!h]
    \centering
    \includegraphics[width=\textwidth]{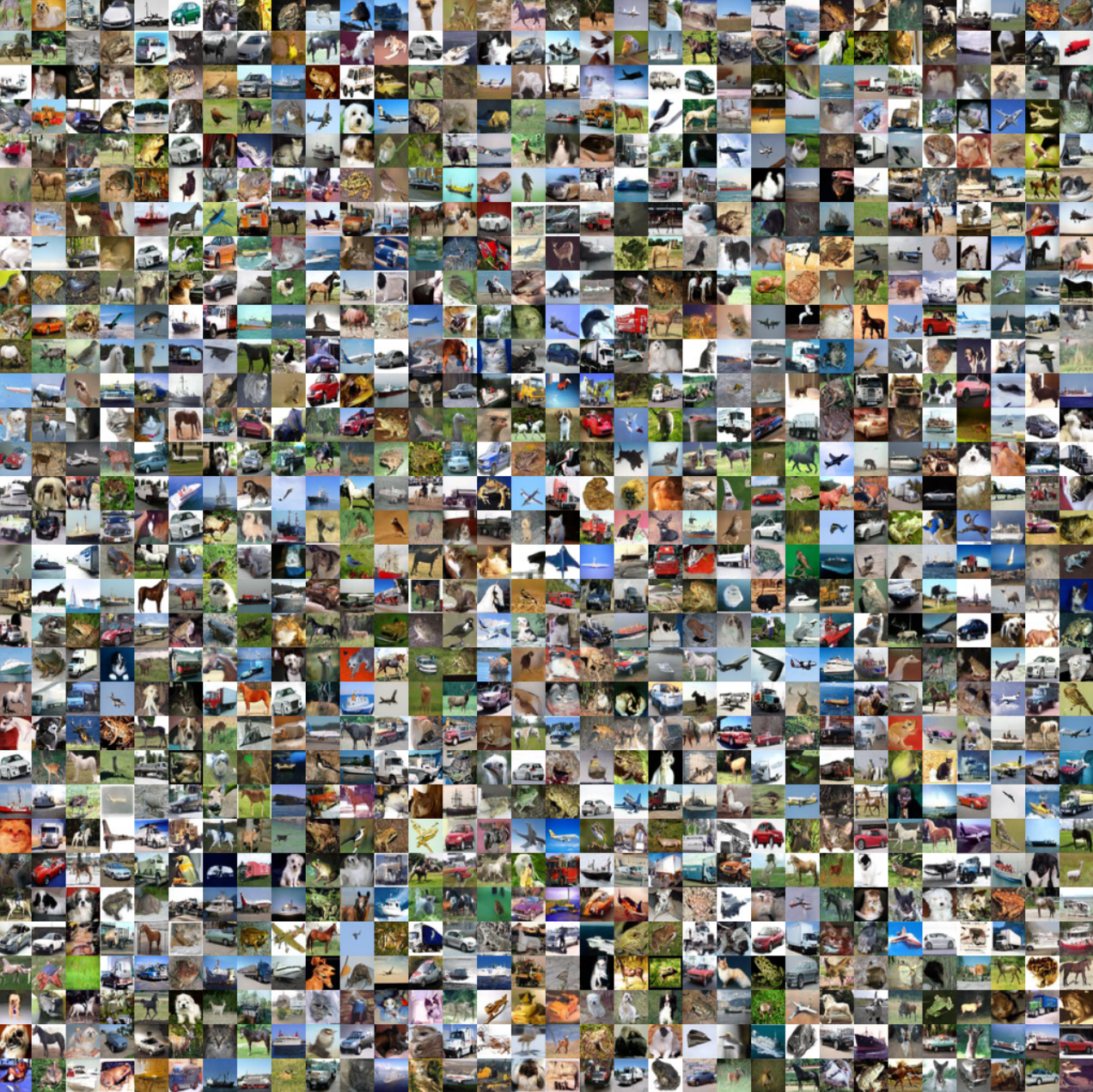}
    \caption{
        One-step samples from our unconditional model on CIFAR-10~(FID=3.77).
        \label{fig:cifar10_uncond}
        }
\end{figure*}

We provide additional qualitative results on ImageNet~(Fig.~\ref{fig:imagenet_cond}), LAION~(Fig.~\ref{fig:more_results},~\ref{fig:laion1},~\ref{fig:laion2},~\ref{fig:laion3}), and CIFAR-10~(Fig.~\ref{fig:cifar10_cond},~\ref{fig:cifar10_uncond}).

\clearpage 

\clearpage 

\clearpage 

\clearpage 

\clearpage 

\clearpage 

\clearpage

\end{document}

%% file: preamble.tex
\usepackage[dvipsnames]{xcolor}

\newif\ifdraft
\drafttrue

\ifdraft
    \newcommand{\TP}[1]{\textcolor{purple}{TP: #1}}
    \newcommand{\MG}[1]{\textcolor{teal}{MG: #1}}
    \newcommand{\fredo}[1]{\textcolor{red}{FD: #1}}
    \newcommand{\bill}[1]{\textcolor{blue}{Bill: #1}}    
    \newcommand{\ES}[1]{\textcolor{olive}{ES: #1}}
    \newcommand{\TY}[1]{
    \textcolor{blue}{TY: #1}
    }

\else
    \newcommand{\TP}[1]{}
    \newcommand{\MG}[1]{}
    \newcommand{\fredo}[1]{}
    \newcommand{\bill}[1]{}    \newcommand{\ES}[1]{}
    \newcommand{\TY}[1]{}
\fi

\usepackage[ruled, lined, linesnumbered, commentsnumbered, longend]{algorithm2e}
\usepackage{listings}
\usepackage{graphicx}
\usepackage{amsmath}
\usepackage{amssymb}
\usepackage{booktabs}
\usepackage{array}
\usepackage{amsmath}
\usepackage{amsfonts} 
\usepackage{graphicx}
\usepackage{caption}
\usepackage[export]{adjustbox}
\usepackage{makecell}
\usepackage{multirow}
\usepackage{rotating}
\usepackage{environ}
\usepackage{enumerate}

\DeclareMathOperator*{\expect}{\mathbb{E}}

\NewEnviron{Answer}
{%
\rotatebox[origin=c]{180}{%
\begin{minipage}[t]
\BODY
\end{minipage}%
}%
}%

%% file: 00_abstract.tex
\begin{abstract}
\vspace{-0.15in}
Diffusion models generate high-quality images but require dozens of forward passes.
We introduce Distribution Matching Distillation (DMD), a procedure to transform a diffusion model into a one-step image generator with minimal impact on image quality.
We enforce the one-step image generator match the diffusion model at distribution level, by minimizing an approximate KL divergence whose gradient can be expressed as the difference between 2 score functions, one of the target distribution and the other of the synthetic distribution being produced by our one-step generator. The score functions are parameterized as two diffusion models trained separately on each distribution.
Combined with a simple regression loss matching the large-scale structure of the multi-step diffusion outputs, our method outperforms all published few-step diffusion approaches, reaching 2.62 FID on ImageNet 64$\times$64 and 11.49 FID on zero-shot COCO-30k, comparable to Stable Diffusion but orders of magnitude faster.
Utilizing FP16 inference, our model can generate images at 20 FPS on modern hardware.
\end{abstract}

%% file: 01_intro.tex
Diffusion models~\cite{ramesh2022hierarchical, rombach2022high, saharia2022photorealistic, song2020score, ho2020denoising, sohl2015deep} have revolutionized image generation, achieving unprecedented levels of realism and diversity with a stable training procedure.
In contrast to GANs~\cite{Goodfellow2014GAN} and VAEs~\cite{kingma2013auto}, however, their sampling is a slow, iterative process that transforms a Gaussian noise sample into an intricate image by progressive denoising~\cite{song2020score, ho2020denoising}. 
This typically requires tens to hundreds of costly neural network evaluations, limiting interactivity in using the generation pipeline as a creative tool.

To accelerate sampling speed, previous methods~\cite{luhman2021knowledge, zheng2022fast, zhao2023unipc, liu2022flow, liu2023instaflow, song2023consistency, salimans2022progressive, meng2023distillation, luo2023latent} distill the noise$\rightarrow$image mapping, discovered by the original multi-step diffusion sampling, into a single-pass student network.
However, fitting such a high-dimensional, complex mapping is certainly a demanding task. 
A challenge is the expensive cost of running the full denoising trajectory, just to realize one loss computation of the student model. 
Recent methods mitigate this by progressively increasing the sampling distance of the student, without running the full denoising sequence of the original diffusion~\cite{salimans2022progressive,meng2023distillation,liu2022flow, liu2023instaflow,song2023consistency,berthelot2023tract,gu2023boot}.
However, the performance of distilled models still lags behind the original multi-step diffusion model.

In contrast, rather than enforcing correspondences between noise and diffusion-generated images, we simply enforce that the student generations look indistinguishable from the original diffusion model. At high level, our goal shares motivation with other \textit{distribution-matching} generative models, such as GMMN~\cite{li2015generative} or GANs~\cite{Goodfellow2014GAN}. Still, despite their impressive success in creating realistic images~\cite{karras2017progressive,karras2020analyzing}, scaling up the model on the general text-to-image data has been challenging~\cite{reed2016generative, zhang2018stackgan++,kang2023scaling}.
In this work, we bypass the issue by starting with a diffusion model that is already trained on large-scale text-to-image data. 
Concretely, we finetune the pretrained diffusion model to learn not only the data distribution, but also the \textit{fake} distribution that is being produced by our distilled generator.
Since diffusion models are known to approximate the score functions on diffused distributions~\cite{hyvarinen2005estimation,song2019generative}, we can interpret the denoised diffusion outputs as gradient directions for making an image ``more realistic", or if the diffusion model is learned on the fake images, ``more fake".
Finally, the gradient update rule for the generator is concocted as the difference of the two, nudging the synthetic images toward higher realism and lower fakeness. Previous work~\cite{wang2023prolificdreamer}, in a method called Variational Score Distillation, shows that modeling the real and fake distributions with a pretrained diffusion model is also effective for test-time optimization of 3D objects. Our insight is that a similar approach can instead train \textit{an entire generative model}.

Furthermore, we find that pre-computing a modest number of the multi-step diffusion sampling outcomes and enforcing a simple regression loss with respect to our one-step generation serves as an effective regularizer in the presence of the distribution matching loss.
Moreover, the regression loss ensures our one-step generator aligns with the teacher model~(see Figure~\ref{fig:results}), demonstrating potential for real-time design previews. 
Our method draws upon inspiration and insights from VSD~\cite{wang2023prolificdreamer}, GANs~\cite{Goodfellow2014GAN}, and pix2pix~\cite{isola2017image}, showing that by (1) modeling real and fake distributions with diffusion models and (2) using a simple regression loss to match the multi-step diffusion outputs, we can train a one-step generative model with high fidelity. %

We evaluate models trained with our Distribution Matching Distillation procedure (\method) across various tasks, including image generation on CIFAR-10~\cite{krizhevsky2009learning} and ImageNet 64$\times$64~\cite{deng2009imagenet}, and zero-shot text-to-image generation on MS COCO 512$\times$512~\cite{lin2014microsoft}.
On all benchmarks, our one-step generator significantly outperforms all published few-steps diffusion methods, such as Progressive Distillation~\cite{salimans2022progressive, meng2023distillation}, Rectified Flow~\cite{liu2022flow, liu2023instaflow}, and Consistency Models~\cite{song2023consistency, luo2023latent}. 
On ImageNet, \method reaches FIDs of 2.62, an improvement of $2.4\times$ over Consistency Model~\cite{song2023consistency}. 
Employing the identical denoiser architecture as Stable Diffusion~\cite{rombach2022high}, \method achieves a competitive FID of 11.49 on MS-COCO 2014-30k. 
Our quantitative and qualitative evaluations show that the images generated by our model closely resemble the quality of those generated by the costly Stable Diffusion model. 
Importantly, our approach maintains this level of image fidelity while achieving a $100\times$ reduction in neural network evaluations.
This efficiency allows \method to generate $512\times512$ images at a rate of 20 FPS when utilizing FP16 inference, opening up a wide range of possibilities for interactive applications.

%% file: 02_related.tex
\noindent 
\textbf{Diffusion Model}
Diffusion models~\cite{song2020score, ho2020denoising, sohl2015deep, balaji2022ediffi} have emerged as a powerful generative modeling framework, achieving unparalleled success in diverse domains such as image generation~\cite{rombach2022high, ramesh2022hierarchical, saharia2022photorealistic}, audio synthesis~\cite{chen2020wavegrad, kong2020diffwave}, and video generation~\cite{ho2022imagen, singer2022make, esser2023structure}.
These models operate by progressively transforming noise into coherent structures through a reverse diffusion process~\cite{song2020denoising, song2020score}. 
Despite state-of-the-art results,
the inherently iterative procedure of diffusion models entails a high and often prohibitive computational cost for real-time applications.
Our work builds upon leading diffusion models~\cite{karras2022elucidating, rombach2022high} and introduces a simple distillation pipeline that reduces the multi-step generative process to a single forward pass.
Our method is universally applicable to any diffusion model with deterministic sampling~\cite{karras2022elucidating, song2020denoising, song2020score}. \\

\begin{figure*}[!t]
    \vspace{0mm}
    \centering
    \includegraphics[width=\linewidth]{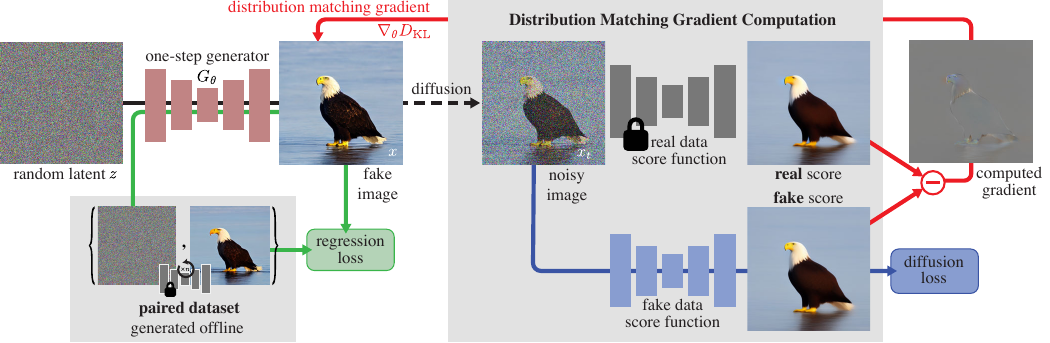}
    \caption{
        {\bf Method overview.} We train one-step generator $G_\theta$ to map random noise $z$ into a realistic image. To match the multi-step sampling outputs of the diffusion model, we pre-compute a collection of noise--image pairs, and occasionally load the noise from the collection and enforce LPIPS~\cite{zhang2018unreasonable} \colorbox{regressionloss}{regression loss} between our one-step generator and the diffusion output. Furthermore, we provide \textbf{\textcolor{distrmatchgrad}{distribution matching gradient $\nabla_\theta D_{KL}$}} to the fake image to enhance realism. We inject a random amount of noise to the fake image and pass it to two diffusion models, one pretrained on the real data and the other continually trained on the fake images with a \colorbox{diffusionloss}{diffusion loss}, to obtain its denoised versions. The denoising scores~(visualized as mean prediction in the plot) indicate directions to make the images more realistic or fake. The difference between the two represents the direction toward more realism and less fakeness and is backpropagated to the one-step generator.
        }
        \label{fig:main}
    \vspace{0mm}
\end{figure*}

\noindent 
\textbf{Diffusion Acceleration}
Accelerating the inference process of diffusion models has been a key focus in the field, leading to the development of two types of approaches. 
The first type advances fast diffusion samplers~\cite{lu2022dpm, lu2022dpm++, karras2022elucidating, liu2022pseudo, zhao2023unipc}, which can dramatically reduce the number of sampling steps required by pre-trained diffusion models—from a thousand down to merely 20-50. However, a further reduction in steps often results in a catastrophic decrease in performance.
Alternatively, diffusion distillation has emerged as a promising avenue for further boosting speed~\cite{song2023consistency, liu2022flow, salimans2022progressive, meng2023distillation, gu2023boot, berthelot2023tract, zheng2022fast, luhman2021knowledge, xiao2021tackling}. 
They frame diffusion distillation as knowledge distillation~\cite{hinton2015distilling}, where a student model is trained to distill the multi-step outputs of the original diffusion model into a single step. Luhman \etal~\cite{luhman2021knowledge} and DSNO~\cite{zheng2023fast} proposed a simple approach of pre-computing the denoising trajectories and training the student model with a regression loss in pixel space. However, a significant challenge is the expensive cost of running the full denoising trajectory for each realization of the loss function. To address this issue, Progressive Distillation (PD)~\cite{salimans2022progressive,meng2023distillation} train a series of student models that halve the number of sampling steps of the previous model. InstaFlow~\cite{liu2022flow,liu2023instaflow} progressively learn straighter flows on which the one step prediction maintains accuracy over a larger distance. Consistency Distillation~(CD)~\cite{song2023consistency}, TRACT~\cite{berthelot2023tract}, and BOOT~\cite{gu2023boot} train a student model to match its own output at a different timestep on the ODE flow, which in turn is enforced to match its own output at yet another timestep. 
In contrast, our method shows that the simple approach of Luhman~\etal and DSNO to pre-compute the diffusion outputs is sufficient, once we introduce distribution matching as the training objective. \\

\noindent 
\textbf{Distribution Matching}
Recently, a few classes of generative models have shown success in scaling up to complex datasets by recovering samples that are corrupted by a predefined mechanism, such as noise injection~\cite{ho2020denoising,ramesh2022hierarchical,saharia2022photorealistic} or token masking~\cite{ramesh2021zero,yu2022scaling,chang2023muse}. On the other hand, there exist generative methods that do not rely on sample reconstruction as the training objective. Instead, they match the synthetic and target samples at a distribution level, such as GMMD~\cite{li2015generative,dziugaite2015training} or GANs~\cite{Goodfellow2014GAN}. Among them, GANs have shown unprecedented quality in realism~\cite{karras2017progressive, karras2019style, karras2020analyzing, brock2018large, sauer2023stylegan, kang2023scaling}, particularly when the GAN loss can be combined with task-specific, auxiliary regression losses to mitigate training instability, ranging from paired image translation~\cite{isola2017image,wang2018pix2pixHD,park2019semantic,zhao2021large} to unpaired image editing~\cite{zhu2017unpaired,lee2020drit++,park2020swapping}. 
Still, GANs are a less popular choice for text-guided synthesis, as careful architectural design is needed to ensure training stability at large scale~\cite{kang2023scaling}. 

Lately, several works~\cite{yi2023monoflow, asokan2023gans, franceschi2023unifying, weber2023score} drew connections between score-based models and distribution matching.
In particular, ProlificDreamer~\cite{wang2023prolificdreamer} introduced Variational Score Distillation (VSD), which leverages a pretrained text-to-image diffusion model as a distribution matching loss. Since VSD can utilize a large pretrained model for unpaired settings~\cite{poole2022dreamfusion, hertz2023delta}, it showed impressive results at particle-based optimization for text-conditioned 3D synthesis. 
Our method refines and extends VSD for training a deep generative neural network for distilling diffusion models. 
Furthermore, motivated by the success of GANs in image translation, we complement the stability of training with a regression loss. As a result, our method successfully attains high realism on a complex dataset like LAION~\cite{schuhmann2022laion}. Our method is different from recent works that combine GANs with diffusion~\cite{xiao2021tackling,wang2022diffusion, xu2023ufogen, sauer2023adversarial}, as our formulation is not grounded in GANs. 
Our method shares motivation with concurrent works~\cite{ye2023score,luo2023diff} that leverage the VSD objective to train a generator, but differs in that we specialize the method for diffusion distillation by introducing regression loss and showing state-of-the-art results for text-to-image tasks.

%% file: 04_method.tex
Our goal is to distill a given pretrained diffusion denoiser, the \emph{base model}, $\mu_\text{base}$, into a fast ``one-step'' image generator, $G_\theta$, that produces high-quality images without the costly iterative sampling procedure (Sec.~\ref{sec:onestep-gen}). While we wish to produce samples from the same distribution, we do not necessarily seek to reproduce the exact mapping.

By analogy with GANs, we denote the outputs of the distilled model as \emph{fake}, as opposed to the \emph{real} images from the training  distribution.
We illustrate our approach in Figure~\ref{fig:main}. We train the fast generator by minimizing the sum of two losses: a distribution matching objective (Sec.~\ref{sec:distribution-matching}), whose gradient update can be expressed as the difference of two score functions, and a regression loss (Sec.~\ref{sec:direct-distillation}) that encourages the generator to match the large scale structure of the base model's output on a fixed dataset of noise-image pairs.
Crucially, we use two diffusion denoisers to model the score functions of the real and fake distributions, respectively, perturbed with Gaussian noise of various magnitudes.
Finally, in Section~\ref{sec:cfg}, we show how to adapt our training procedure with classifier-free guidance.

\begin{figure}[!btp]
\centering
\includegraphics[width=\linewidth]{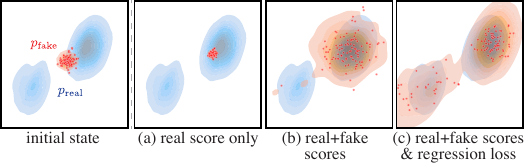}
\caption{\label{fig:various-objectives}
    Optimizing various objectives starting from the same configuration (left) leads to different outcomes.
    (a) Maximizing the real score only, the fake samples all collapse to the closest mode of the real distribution.
    (b) With our distribution matching objective but not regression loss, the generated fake data covers more of the real distribution, but only recovers the closest mode, missing the second mode entirely.
    (c) Our full objective, with the regression loss, recovers both modes of the target distribution.
}   
\end{figure}

\subsection{Pretrained base model and One-step generator}\label{sec:onestep-gen}
Our distillation procedure assumes a pretrained diffusion model $\mu_\text{base}$ is given.
Diffusion models are trained to reverse a Gaussian diffusion process that progressively adds noise to a sample from a real data distribution $x_0 \sim p_\text{real}$, turning it into white noise $x_T\sim\mathcal{N}(0, \mathbf{I})$ over $T$ time steps~\cite{sohl2015deep,ho2020denoising, song2020score}; we use $T=1000$.
We denote the diffusion model as $\mu_\text{base}(x_t, t)$.
Starting from a Gaussian sample $x_T$, the model iteratively denoises a running noisy estimate $x_t$, conditioned on the timestep $t\in \{0, 1, ..., T-1 \}$ (or noise level), to produce a sample of the target data distribution.
Diffusion models typically require 10 to 100s steps to produce realistic images. 
Our derivation uses the mean-prediction form of diffusion for simplicity~\cite{karras2022elucidating} but works identically with $\epsilon$-prediction~\cite{ho2020denoising, rombach2022high} with a change of variable~\cite{kingma2021variational} (see Appendix~\ref{sec:change_of_variable}).
Our implementation uses pretrained models from EDM~\cite{karras2022elucidating} and Stable Diffusion~\cite{rombach2022high}.

\noindent 
\textbf{One-step generator.}
Our one-step generator $G_\theta$ has the architecture of the base diffusion denoiser but without time-conditioning.
We initialize its parameters $\theta$ with the base model,
i.e., $G_{\theta}(z) = \mu_\text{base}(z, T-1), \forall z$, before training. %

\subsection{Distribution Matching Loss}\label{sec:distribution-matching}

Ideally, we would like our fast generator to produce samples that are indistinguishable from real images.
Inspired by the ProlificDreamer~\cite{wang2023prolificdreamer}, we minimize the Kullback–Leibler~(KL) divergence between the real and fake image distributions, $p_\text{real}$ and $p_\text{fake}$, respectively:
\begin{equation}
\small 
    \begin{aligned}
        D_{KL}\left(p_{\text{fake}} \; \| \; p_{\text{real}} \right) &= \expect_{x\sim p_\text{fake}}\left(\log\left(\frac{p_\text{fake}(x)}{p_\text{real}(x)}\right)\right)\\
        &= \expect_{\substack{
        z \sim \mathcal{N}(0; \mathbf{I}) \\
        x = G_\theta(z)
        }}-\big(\log~p_\text{real}(x) - \log~p_\text{fake}(x)\big).
    \end{aligned}
    \label{eq:kl}
\end{equation}
Computing the probability densities to estimate this loss is generally intractable, but we only need the gradient with respect to $\theta$ to train our generator by gradient descent.

\noindent \textbf{Gradient update using approximate scores.}
Taking the gradient of Eq.~\eqref{eq:kl} with respect to the generator parameters:
\vspace{-2mm}
\begin{equation}
\small 
    \begin{aligned}
       \nabla_\theta D_{KL}
        &= \expect_{\substack{
        z \sim \mathcal{N}(0; \mathbf{I}) \\
        x = G_\theta(z)
        } } \Big[-
        \big(
        s_\text{real}(x) - s_\text{fake}(x)\big)
        \hspace{.5mm} \frac{dG}{d\theta}
         \Big],
    \end{aligned}
    \label{eq:kl-grad}
\end{equation}
where $s_\text{real}(x) =\nabla_{x} \text{log}~p_\text{real}(x)$, $s_\text{fake}(x) =\nabla_{x} \text{log}~p_\text{fake}(x)$ are the scores of the respective distributions. Intuitively, $s_\text{real}$ moves $x$ toward the modes of $p_\text{real}$, and $-s_\text{fake}$ spreads them apart, as shown in Figure~\ref{fig:various-objectives}(a, b).
Computing this gradient is still challenging for two reasons:
first, the scores diverge for samples with low probability --- in particular $p_\text{real}$ vanishes for fake samples,
and second, our intended tool for estimating score, namely the diffusion models, only provide scores of the diffused distribution. 
Score-SDE~\cite{song2020score, song2019generative} provides an answer to these two issues.

\begin{figure}[!t]
\centering
\includegraphics[width=\linewidth]{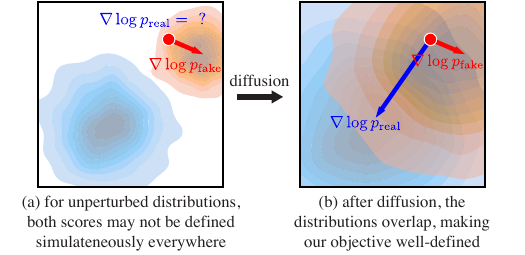}
\caption{\label{fig:diffused-match}
    Without perturbation, the real/fake distributions may not overlap (a).
    Real samples only get a valid gradient from the real score, and fake samples from the fake score.
    After diffusion (b), our distribution matching objective is well-defined everywhere.
}
\vspace{-2mm}
\end{figure}

By perturbing the data distribution with random Gaussian noise of varying standard deviations, we create a family of ``blurred'' distributions that are fully-supported over the ambient space, and therefore overlap, so that the gradient in Eq.~\eqref{eq:kl-grad} is well-defined (Figure~\ref{fig:diffused-match}).
Score-SDE then shows that a trained diffusion model approximates the score function of the diffused distribution.

Accordingly, our strategy is to use a pair of diffusion denoisers to model the scores of the real and fake distributions after Gaussian diffusion. With slight abuse of notation, we define these as $s_\text{real}(x_t,t)$ and $s_{\text{fake}}(x_t,t)$, respectively.
Diffused sample $x_t\sim q(x_t|x)$ is obtained by adding noise to generator output $x=G_\theta(z)$ at diffusion time step $t$:
\begin{equation}
    \label{eq:forward-diffusion}
    q_t(x_t|x)\sim\mathcal{N}(\alpha_t x; \sigma_t^2 \mathbf{I}),
\end{equation}

\noindent where $\alpha_t$ and $\sigma_t$ are from the diffusion noise schedule.

\noindent \textbf{Real score.} %
The real distribution is fixed, corresponding to the training images of the base diffusion model, so we model its score using a fixed copy of the pretrained diffusion model $\mu_\text{base}(x, t)$. 
The score given a diffusion model is given by Song \etal~\cite{song2020score}:  %
\begin{equation}
\label{eq:real-score}
    s_\text{real}(x_t, t) = - \hspace{.5mm} \frac{x_t-\alpha_t\mu_\text{base}(x_t, t)}{\sigma_t^2}.
\end{equation}

\noindent \textbf{Dynamically-learned fake score.}
We derive the fake score function, in the same manner as the real score case:
\begin{equation}
\label{eq:fake-score}
    s_{\text{fake}}(x_t, t) = - \hspace{.5mm}\frac{x_t-\alpha_t\mu_{\text{fake}}^\phi(x_t, t)}{\sigma_t^2}.
\end{equation}

\noindent However, as the distribution of our generated samples changes throughout training, we dynamically adjust the fake diffusion model $\mu_{\text{fake}}^\phi$ to track these changes.
We initialize the fake diffusion model from the pretrained diffusion model $\mu_\text{base}$, updating parameters $\phi$ during training, by minimizing a standard denoising objective~\cite{vincent2011connection, ho2020denoising}:
\begin{equation}
\label{eq:fake_train}
    \mathcal{L}_\text{denoise}^\phi = ||\mu_{\text{fake}}^\phi(x_t, t) - x_0||_2^2,
\end{equation}

\noindent where $\mathcal{L}_\text{denoise}^\phi$ is weighted according to the diffusion timestep $t$, using the same weighting strategy employed during the training of the base diffusion model~\cite{karras2022elucidating, rombach2022high}.

\noindent \textbf{Distribution matching gradient update.}
\label{sec:dm_grad} 
Our final approximate distribution matching gradient is obtained by replacing the exact score in Eq.~\eqref{eq:kl-grad} with those defined by the two diffusion models on the perturbed samples $x_t$ and taking the expectation over the diffusion time steps: %
\begin{equation}
\footnotesize 
\label{eq:dm-grad}
\begin{aligned}
    \nabla_{\theta} D_{KL} &\simeq 
        \expect_{\substack{z, t, x, x_t
        }}
    \left[
        w_t
        \alpha_t
        \big(s_{\text{fake}}(x_t, t) - s_\text{real}(x_t, t)\big)
        \hspace{.5mm} \frac{dG}{d\theta}
        \right], \\
\end{aligned}
\end{equation}
where $z \sim \mathcal{N}(0;\mathbf{I})$,  $x = G_\theta(z)$,
$t \sim \mathcal{U}(T_\text{min}, T_\text{max})$,
and $x_t \sim q_t(x_t|x)$.
We include the derivations in Appendix~\ref{sec:proof_dmd}.
Here, $w_t$ is a time-dependent scalar weight we add to improve the training dynamics.
We design the weighting factor to normalize the gradient's magnitude across different noise levels. 
Specifically, we compute the mean absolute error across spatial and channel dimensions between the denoised image and the input, setting

\begin{equation}
    w_t = \tfrac{\sigma_t^2}{\alpha_t} \tfrac{CS}{|| \mu_\text{base}(x_t, t)-x ||_1},
    \label{eq:weighting}
\end{equation}

\noindent where $S$ is the number of spatial locations and $C$ is the number of channels. In Sec.~\ref{sec:ablation}, we show that this weighting outperforms previous designs~\cite{poole2022dreamfusion, wang2023prolificdreamer}.
We set $T_\text{min} = 0.02\hspace{.5mm}T$ and $T_\text{max}=0.98\hspace{.5mm}T$, following DreamFusion~\cite{poole2022dreamfusion}.

\subsection{Regression loss and final objective}\label{sec:direct-distillation}

The distribution matching objective introduced in the previous section is well-defined for $t\gg0$, i.e., when the generated samples are corrupted with a large amount of noise. However, for a small amount of noise, $s_\text{real}(x_t, t)$ often becomes unreliable, as $p_\text{real}(x_t, t)$ goes to zero. Furthermore, as the score $\nabla_{x} \text{log} (p)$ is invariant to scaling of probability density function $p$, the optimization is susceptible to mode collapse/dropping, where the fake distribution assigns higher overall density to a subset of the modes. To avoid this, we use an additional regression loss to ensure all modes are preserved; see Figure~\ref{fig:various-objectives}(b), (c).

This loss measures the pointwise distance between the generator and base diffusion model outputs, given the \emph{same} input noise.
Concretely, we build a paired dataset $\mathcal{D}=\{z, y\}$ of random Gaussian noise images $z$ and the corresponding outputs $y$, obtained by sampling the pretrained diffusion model $\mu_\text{base}$ using a deterministic ODE solver~\cite{song2020denoising, karras2022elucidating, liu2022pseudo}. 
In our CIFAR-10 and ImageNet experiments, we utilize the Heun solver from EDM~\cite{karras2022elucidating}, with 18 steps for CIFAR-10 and 256 steps for ImageNet.
For the LAION experiments, we use the PNDM~\cite{liu2022pseudo} solver with 50 sampling steps. 
We find that even a small number of noise--image pairs, generated using less than 1\% of the training compute, in the case of CIFAR10, for example, acts as an effective regularizer.
Our regression loss is given by:
\begin{equation}
\begin{split}
    \mathcal{L}_\text{reg}
    &= \expect_{(z, y)\sim \mathcal{D}} \ell (G_\theta(z), y).
\end{split}
\label{eq:distillation}
\end{equation}
\noindent 
We use Learned Perceptual Image Patch Similarity (LPIPS)~\cite{zhang2018unreasonable} as the distance function $\ell$, following InstaFlow~\cite{liu2023instaflow} and Consistency Models~\cite{song2023consistency}. 

\noindent \textbf{Final objective.} Network $\mu_\text{fake}^\phi$ is trained with $\mathcal{L}_\text{denoise}^\phi$, which is used to help calculate $\nabla_{\theta} D_{KL}$.
For training $G_\theta$, the final objective is $D_{KL} + \lambda_\text{reg} \mathcal{L}_\text{reg}$,
using $\lambda_\text{reg}=0.25$ unless otherwise specified. The gradient $\nabla_{\theta} D_{KL}$ is computed in Eq.~\eqref{eq:dm-grad}, and gradient $\nabla_{\theta} \mathcal{L}_\text{reg}$ is computed from Eq.~\eqref{eq:distillation} with automatic differentiation.
We apply the two losses to distinct data streams: unpaired fake samples for the distribution matching gradient and paired examples described in Section~\ref{sec:direct-distillation} for the regression loss.
Algorithm~\ref{alg:distillation} outlines the final training procedure.
Additional details are provided in Appendix~\ref{sec:implementation}.

\begin{algorithm}
    \scriptsize
    \caption{\label{alg:distillation}DMD Training procedure}
    \KwIn{Pretrained real diffusion model $\mu_\text{real}$,
    paired dataset $\mathcal{D}=\{z_\text{ref}, y_\text{ref}\}$
    }
    \KwOut{Trained generator $G$.}
    
    \tcp{Initialize generator and fake score estimators from pretrained model}
    $G \leftarrow \text{copyWeights}(\mu_\text{real}),$
    $\mu_\text{fake} \leftarrow \text{copyWeights}(\mu_\text{real})$

    \While{train}{
        \tcp{Generate images}
        Sample batch $z \sim \mathcal{N}(0, \mathbf{I})^B$ and $(z_\text{ref}, y_\text{ref}) \sim \mathcal{D}$
        
        $x \leftarrow G(z),~x_\text{ref} \leftarrow G(z_\text{ref})$
        
        $x = \text{concat}(x, x_\text{ref}) \text{ \textbf{if} dataset is LAION \textbf{else} } x $
        
        \text{~}
        
        \tcp{Update generator}
        $\mathcal{L}_\text{KL} \leftarrow \text{distributionMatchingLoss}(\mu_\text{real}, \mu_\text{fake}, x)$ \tcp{Eq~\ref{eq:dm-grad}}
        
        $\mathcal{L}_\text{reg} \leftarrow \text{LPIPS}(x_\text{ref}, y_\text{ref})$ \tcp{Eq~\ref{eq:distillation}}
        
        $\mathcal{L}_{G} \leftarrow \mathcal{L}_\text{KL} + \lambda_\text{reg}\mathcal{L}_\text{reg}$
        
        $G \leftarrow \text{update}(G, \mathcal{L}_G)$
        
        \text{~}
        
        \tcp{Update fake score estimation model}
        Sample time step $t\sim\mathcal{U}(0,1)$
        
        $x_t \leftarrow \text{forwardDiffusion}(\text{stopgrad}(x), t)$
        
        $ \mathcal{L}_{\text{denoise}} \leftarrow \text{denoisingLoss}(\mu_\text{fake}(x_t, t), \text{stopgrad}(x))$ \tcp{Eq~\ref{eq:fake_train}}
        
        $\mu_\text{fake} \leftarrow \text{update}(\mu_\text{fake}, \mathcal{L}_{\text{denoise}})$
    }
\end{algorithm}

\subsection{Distillation with classifier-free guidance}\label{sec:cfg}

Classifier-Free Guidance~\cite{ho2022classifier} is widely used to improve the image quality of text-to-image diffusion models.
Our approach also applies to diffusion models that use classifier-free guidance.
We first generate the corresponding noise-output pairs by sampling from the guided model to construct the paired dataset needed for regression loss $\mathcal{L}_\text{reg}$.
When computing the distribution matching gradient $\nabla_{\theta} D_{KL}$,
we substitute the real score with that derived from the mean prediction of the guided model.
Meanwhile, we do not modify the formulation for the fake score.
We train our one-step generator with a fixed guidance scale. 

%% file: 05_experiments.tex
We assess the capabilities of our approach using several benchmarks, including class-conditional generation on CIFAR-10~\cite{krizhevsky2009learning} and ImageNet~\cite{deng2009imagenet}. 
We use the Fréchet Inception Distance (FID)~\cite{heusel2017gans} to measure image quality and CLIP Score~\cite{radford2021learning} to evaluate text-to-image alignment.
First, we perform a direct comparison on ImageNet (Sec.~\ref{sec:class_image}), where our distribution matching distillation substantially outperforms competing distillation methods with identical base diffusion models. 
Second, we perform detailed ablation studies verifying the effectiveness of our proposed modules~(Sec.~\ref{sec:ablation}).
Third, we train a text-to-image model on the LAION-Aesthetic-6.25+ dataset~\cite{schuhmann2022laion} with a classifier-free guidance scale of 3~(Sec.~\ref{sec:text_image}).
In this phase, we distill Stable Diffusion v1.5, and we show that our distilled model achieves FID comparable to the original model, while offering a 30$\times$ speed-up.
Finally, we train another text-to-image model on LAION-Aesthetic-6+, utilizing a higher guidance value of 8~(Sec.~\ref{sec:text_image}). 
This model is tailored to enhance visual quality rather than optimize the FID metric. 
Quantitative and qualitative analysis confirm that models trained with our distribution matching distillation procedure can produce high-quality images rivaling Stable Diffusion. 
We describe additional training and evaluation details in the appendix.

\subsection{Class-conditional Image Generation}\label{sec:class_image}
We train our model on class-conditional ImageNet-64×64 and benchmark its performance with competing methods.
Results are shown in Table~\ref{table:imagenet}. 
Our model surpasses established GANs like BigGAN-deep~\cite{brock2018large} and recent diffusion distillation methods, including the Consistency Model~\cite{song2023consistency} and TRACT~\cite{berthelot2023tract}. 
Our method remarkably bridges the fidelity gap, achieving a near-identical FID score (within 0.3) compared to the original diffusion model, while also attaining a 512-fold increase in speed.
On CIFAR-10, our class-conditional model reaches a competitive FID of 2.66. 
We include the CIFAR-10 results in the appendix.

\begin{table}[h]
\centering
\small 
\begin{tabular}{lrr}
\toprule
\multirow{2}{*}{Method} & \multirow{2}{*}{\shortstack[c]{\# Fwd \\ Pass~($\downarrow$)}} & \multirow{2}{*}{\shortstack[c]{FID \\ ($\downarrow$)}} \\
\\
\midrule  
BigGAN-deep
\cite{brock2018large}
& 1 & 4.06 \\
ADM
\cite{dhariwal2021diffusion}
& 250 & \textbf{2.07}  \\
\midrule 
Progressive Distillation \cite{salimans2022progressive} & 1 & 15.39 \\
DFNO \cite{zheng2022fast} & 1 & 7.83  \\
BOOT~\cite{gu2023boot} & 1 & 16.30  \\
TRACT~\cite{berthelot2023tract} & 1 & 7.43 \\ 
Meng et al.~\cite{meng2023distillation} & 1 & 7.54  \\
Diff-Instruct~\cite{luo2023diff} & 1 & 5.57 \\ 
Consistency Model~\cite{song2023consistency} & 1 & 6.20\\
\textbf{\method~(Ours)} & 1 & \textbf{2.62}\\
\midrule 
EDM$^\dagger$ (Teacher)
\cite{karras2022elucidating}
& 512 & 2.32  \\
\bottomrule
\end{tabular}
\caption{
\label{table:imagenet}
Sample quality comparison on ImageNet-64$\times$64. Baseline numbers are derived from Song et al.~\cite{song2023consistency}. The upper section of the table highlights popular diffusion and GAN approaches~\cite{brock2018large, dhariwal2021diffusion}.
The middle section includes a list of competing diffusion distillation methods. 
The last row shows the performance of our teacher model, EDM$^\dagger$ \cite{karras2022elucidating}.}
\end{table}

\subsection{Ablation Studies}
\label{sec:ablation}
\begin{figure}[!tbp]
\begin{subfigure}[b]{\linewidth}
\centering
\includegraphics[height=0.5\linewidth,keepaspectratio]{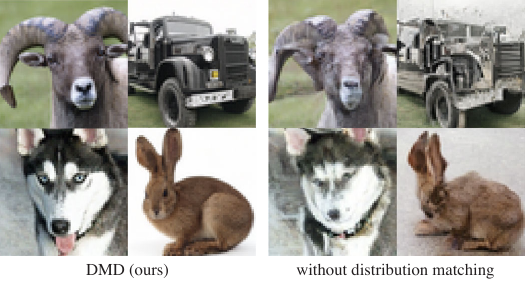}
\caption{Qualitative comparison between our model (\emph{left}) and the baseline model excluding the distribution matching objective (\emph{right}). The baseline model generates images with compromised realism and structural integrity. Images are generated from the same random seed.}
\end{subfigure}

\vspace{2mm}

\begin{subfigure}[b]{\linewidth}
\centering
\includegraphics[height=0.5\linewidth,keepaspectratio]{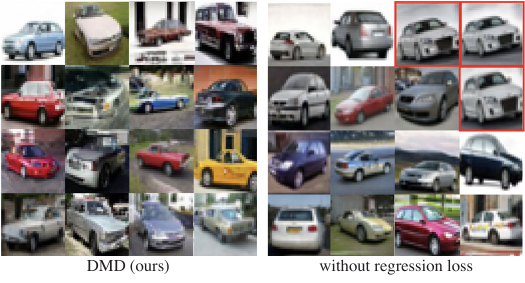}
\caption{Qualitative comparison between our model (\emph{left}) and the baseline model omitting the regression loss (\emph{right}). The baseline model tends to exhibit mode collapse and a lack of diversity, as evidenced by the predominant appearance of the grey car~(highlighted with a red square). Images are generated from the same random seed.
}
\end{subfigure}

\caption{\label{fig:ablation} Ablation studies of our training loss, including the distribution matching objective~(top) and the regression loss~(bottom).}
\vspace{-5mm}
\end{figure}

We first compare our method with two baselines: one omitting the distribution matching objective and the other missing the regression loss in our framework.
Table~\ref{tab:ablations} (left) summarizes the results. 
In the absence of distribution matching loss, our baseline model produces images that lack realism and structural integrity, as illustrated in the top section of Figure \ref{fig:ablation}. 
Likewise, omitting the regression loss leads to training instability and a propensity for mode collapse, resulting in a reduced diversity of the generated images.
This issue is illustrated in the bottom section of Figure \ref{fig:ablation}.

Table~\ref{tab:ablations} (right) demonstrates the advantage of our proposed sample weighting strategy~(Section \ref{sec:dm_grad}).
We compare with $\sigma_t /\alpha_t$ and $\sigma_t^3 /\alpha_t$, two popular weighting schemes utilized by DreamFusion~\cite{poole2022dreamfusion} and ProlificDreamer~\cite{wang2023prolificdreamer}. 
Our weighting strategy achieves a healthy 0.9 FID improvement as it normalizes the gradient magnitudes across noise levels and stabilizes the optimization. 

\begin{table}[h]
  \setlength{\tabcolsep}{1pt}
  \begin{minipage}{0.29\textwidth} %
    \small
    \centering
    \begin{tabular}{lccc}
    \toprule
    Training loss & CIFAR~ & ImageNet~ \\ %
    \midrule
    w/o Dist. Matching & 3.82 & 9.21 \\
    w/o Regress. Loss & 5.58 & 5.61 \\
    \textbf{\method~(Ours)} & \textbf{2.66}  & \textbf{2.62}  \\
    \bottomrule
    \end{tabular}
  \end{minipage}%
  \begin{minipage}{0.2\textwidth} %
    \small
    \centering
    \begin{tabular}{lc}
\toprule
Sample weighting & CIFAR  \\
\midrule
$\sigma_t/\alpha_t$~\cite{poole2022dreamfusion}  & 3.60  \\\
$\sigma_t^3/\alpha_t$~\cite{poole2022dreamfusion, wang2023prolificdreamer} & 3.71  \\
\textbf{Eq. \ref{eq:weighting}~(Ours)}& \textbf{2.66} \\ 
 \bottomrule
\end{tabular}
    \end{minipage}
\caption{{\bf Ablation study.} \textit{(left)} We ablate elements of our training loss. We show the FID results on CIFAR-10 and ImageNet-64$\times$64. \textit{(right)} We compare different sample weighting strategies for the distribution matching loss.}
\label{tab:ablations}
\vspace{-.1in}
\end{table}

\begin{figure*}[!h]
    \centering
    \includegraphics[width=0.94\textwidth]{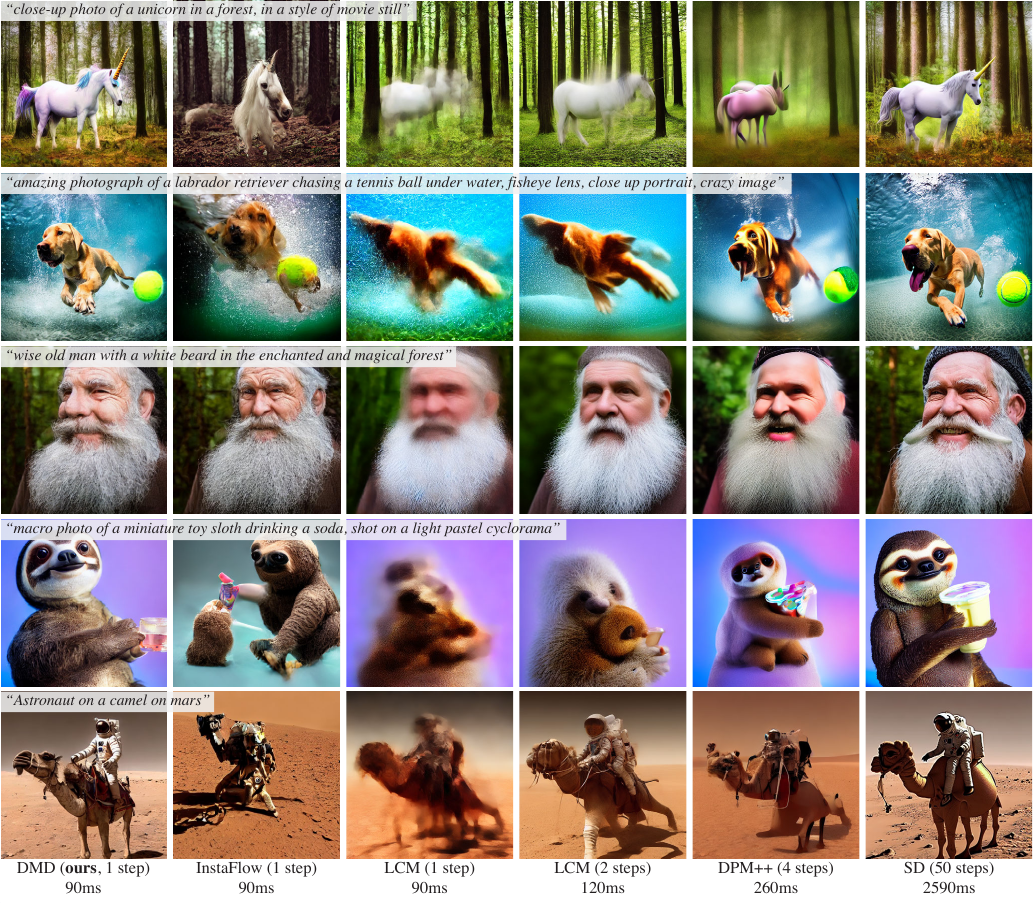}
    \caption{
        Starting from a pretrained diffusion model, here Stable Diffusion (right),
        our distribution matching distillation algorithm yields a model that can generate
        images with much higher quality (left)
        than previous few-steps generators (middle),
        with the same speed or faster. 
        \label{fig:results}
        }
        \vspace{-3mm}
\end{figure*}

\subsection{Text-to-Image Generation}
\label{sec:text_image}
We use zero-shot MS COCO to evaluate our model's performance for text-to-image generation. 
We train a text-to-image model by distilling Stable Diffusion v1.5~\cite{rombach2022high} on the LAION-Aesthetics-6.25+~\cite{schuhmann2022laion}.
We use a guidance scale of 3, which yields the best FID for the base Stable Diffusion model. 
The training takes around 36 hours on a cluster of 72 A100 GPUs. 
Table~\ref{table:laion_guidance3} compares our model to state-of-the-art approaches. 
Our method showcases superior performance over StyleGAN-T~\cite{sauer2023stylegan}, surpasses all other diffusion acceleration methods, including advanced diffusion solvers~\cite{lu2022dpm++, zhao2023unipc}, and diffusion distillation techniques such as Latent Consistency Models~\cite{luo2023latentlora, luo2023latent}, UFOGen~\cite{xu2023ufogen}, and InstaFlow~\cite{liu2023instaflow}.
We substantially close the gap between distilled and base models, reaching within $2.7$ FID from Stable Diffusion v1.5,
while running approximately 30$\times$ faster. 
With FP16 inference, our model generates images at 20 frames per second, enabling interactive applications.

\begin{table}[h]
\centering
\small 
\resizebox{\linewidth}{!}{
\begin{tabular}{ll@{\ }c@{\ }c@{\ }c}
\toprule
Family & Method & Resolution~($\uparrow$) & Latency ($\downarrow$) & FID ($\downarrow$)   \\
\midrule
\multirow{9}{*}{\shortstack[l]{\textbf{Original, }\\ \textbf{unaccelerated}}} &
DALL$\cdot$E~\cite{ramesh2021zero} & 256 & - & 27.5 \\
& DALL$\cdot$E 2~\cite{ramesh2022hierarchical} & 256 & - & 10.39 \\
& Parti-750M~\cite{yu2022scaling} & 256 & - & 10.71  \\ 
& Parti-3B~\cite{yu2022scaling} & 256 & 6.4s & 8.10  \\ 
& Make-A-Scene~\cite{gafni2022make} & 256 & 25.0s & 11.84 \\ 
& GLIDE~\cite{nichol2021glide} & 256 & 15.0s & 12.24 \\ 
& LDM~\cite{rombach2022high} & 256 & 3.7s & 12.63 \\ 
& Imagen~\cite{saharia2022photorealistic} & 256 & 9.1s & 7.27 \\ 
& eDiff-I~\cite{balaji2022ediffi} & 256 & 32.0s & \textbf{6.95} \\
\midrule 
\multirow{3}{*}{\shortstack[l]{\textbf{GANs}}} &
LAFITE~\cite{zhou2022towards} & 256 & 0.02s & 26.94 \\
& StyleGAN-T~\cite{sauer2023stylegan} & 512 & 0.10s & 13.90 \\
& GigaGAN~\cite{kang2023scaling} & 512 & 0.13s & \textbf{9.09} \\
\midrule 
\multirow{7}{*}{\shortstack[l]{\textbf{Accelerated} \\ \textbf{diffusion}}} &
DPM++~(4 step)~\cite{lu2022dpm++}$^\dagger$ & 512 & 0.26s & 22.36  \\
& UniPC~(4 step)~\cite{zhao2023unipc}$^\dagger$ & 512 & 0.26s & 19.57 \\
& LCM-LoRA~(4 step)\cite{luo2023latentlora}$^\dagger$ & 512 & 0.19s & 23.62  \\
& InstaFlow-0.9B~\cite{liu2023instaflow} & 512 & 0.09s & 13.10  \\
& UFOGen~\cite{xu2023ufogen} & 512 & 0.09s & 12.78 \\ 
& \textbf{\method~(Ours)} & 512 & 0.09s & \textbf{11.49} 
\\
\midrule 
\textbf{Teacher} & SDv1.5$^\dagger$~\cite{rombach2022high} & 512 & 2.59s & 8.78  \\
\bottomrule
\end{tabular}
}
\caption{
\label{table:laion_guidance3}
{\bf Sample quality comparison on zero-shot text-to-image generation on MS COCO-30k.}  Baseline numbers are derived from GigaGAN~\cite{kang2023scaling}. The dashed line indicates that the result is unavailable. $^\dagger$Results are evaluated by us using the released models. 
LCM-LoRA is trained with a guidance scale of 7.5. 
We use a guidance scale of 3 for all the other methods. Latency is measured with a batch size of 1.}
\end{table}

\paragraph{High guidance-scale diffusion distillation.}
For text-to-image generation, diffusion models typically operate with a high guidance scale to enhance image quality~\cite{rombach2022high, podell2023sdxl}. 
To evaluate our distillation method in this high guidance-scale regime, we trained an additional text-to-image model. 
This model distills SD v1.5 using a guidance scale of 8 on the LAION-Aesthetics-6+ dataset~\cite{schuhmann2022laion}. 
Table~\ref{table:laion_guidance8} benchmarks our approach against various diffusion acceleration methods~\cite{lu2022dpm++, zhao2023unipc, luo2023latentlora}.
Similar to the low guidance model, our one-step generator significantly outperforms competing methods, even when they utilize a four-step sampling process.
Qualitative comparisons with competing approaches and the base diffusion model are shown in Figure~\ref{fig:results}.

\begin{table}[h]
\centering
\small 
\resizebox{\linewidth}{!}{
\begin{tabular}{l@{\ }c@{\ }c@{\ }c}
\toprule
Method & Latency~($\downarrow$) & FID~($\downarrow$) & CLIP-Score~($\uparrow$) \\
\midrule
DPM++~(4 step)\cite{lu2022dpm++}$^\dagger$ & 0.26s & 22.44 & 0.309 \\
UniPC~(4 step)\cite{zhao2023unipc}$^\dagger$ & 0.26s & 23.30 & 0.308 \\ 
LCM-LoRA~(1 step)~\cite{luo2023latentlora}$^\dagger$ & 0.09s & 77.90 & 0.238 \\
LCM-LoRA~(2 step)~\cite{luo2023latentlora}$^\dagger$ & 0.12s & 24.28 & 0.294 \\
LCM-LoRA~(4 step)~\cite{luo2023latentlora}$^\dagger$ & 0.19s & 23.62 & 0.297\\
\textbf{\method~(Ours)} & 0.09s & \textbf{14.93} & \textbf{0.320} \\
\midrule 
SDv1.5$^\dagger$~(Teacher)~\cite{rombach2022high} & 2.59s & 13.45 & 0.322  \\ 
\bottomrule
\end{tabular}

}
\caption{
\label{table:laion_guidance8}
{\bf FID/CLIP-Score comparison on MS COCO-30K.} $^\dagger$Results are evaluated by us. LCM-LoRA is trained with a guidance scale of 7.5. We use a guidance scale of 8 for all the other methods. Latency is measured with a batch size of 1.
}
\end{table}